\documentclass[10pt,twocolumn,letterpaper]{article}

\usepackage[pagenumbers]{cvpr} % To force page numbers, e.g. for an arXiv version

\usepackage{graphicx}
\usepackage{amsmath}
\usepackage{amssymb}
\usepackage{booktabs}
\usepackage{makecell}

\usepackage{multirow}
\usepackage{caption}
\usepackage{adjustbox}
\usepackage{tabularx}
\usepackage{appendix}

\usepackage{etoolbox}

\usepackage[pagebackref,breaklinks,colorlinks]{hyperref}

\usepackage[capitalize]{cleveref}
\crefname{section}{Sec.}{Secs.}
\Crefname{section}{Section}{Sections}
\Crefname{table}{Table}{Tables}
\crefname{table}{Tab.}{Tabs.}

\newcommand{\Fref}[1]{Figure~\ref{#1}}

\begin{document}

\title{HairCLIP:~Design Your Hair by Text and Reference Image}

\author{ Tianyi Wei\textsuperscript{\rm 1}, Dongdong Chen\textsuperscript{\rm 2}, Wenbo Zhou\textsuperscript{\rm 1}, Jing Liao\textsuperscript{\rm 3},\\
	 Zhentao Tan\textsuperscript{\rm 1},  Lu Yuan\textsuperscript{\rm 2}, Weiming Zhang\textsuperscript{\rm 1}, Nenghai Yu\textsuperscript{\rm 1} \\
	\normalsize\textsuperscript{\rm 1}University of Science and Technology of China  \ \textsuperscript{\rm 2}Microsoft Cloud AI  \   \\ 
	\normalsize\textsuperscript{\rm 3}City University of Hong Kong\    \\
%	{\tt\small\{bestwty@mail., welbeckz@, tzt@mail., zhangwm@, ynh@\}ustc.edu.cn } \\
%	{\tt\small cddlyf@gmail.com}, {\tt\small jingliao@cityu.edu.hk}, {\tt\small luyuan@microsoft.com}
}

\twocolumn[{
\renewcommand\twocolumn[1][]{#1}
\maketitle
    \vspace{-1.9em}
    \setlength\tabcolsep{0.5pt}
    \centering
    \small
    \begin{tabular}{c}
        \includegraphics[width=0.99\textwidth]{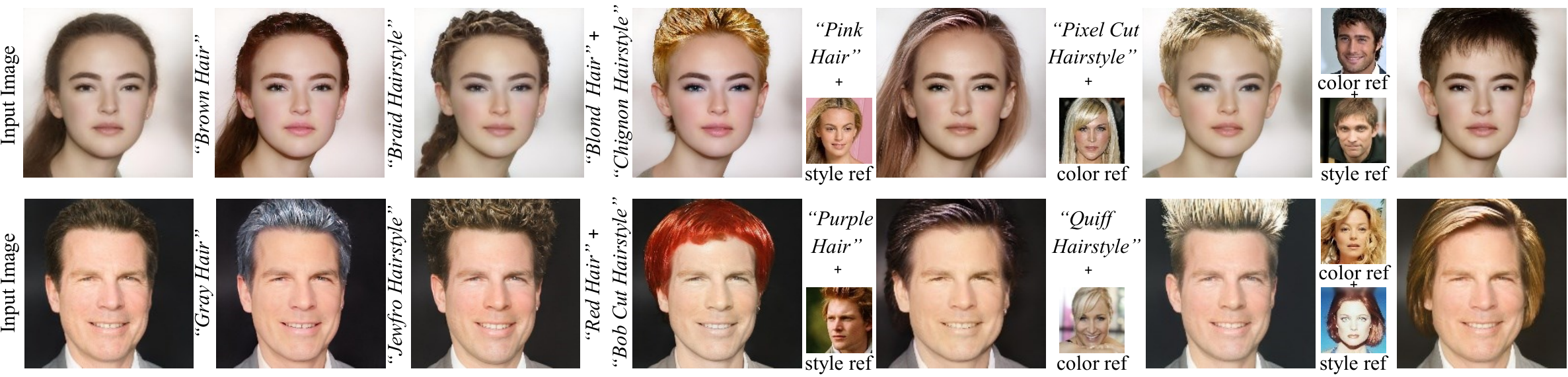}
    \end{tabular}
    \vspace{-0.8em}
    \captionof{figure}{Our single framework supports hairstyle and hair color editing individually or jointly, and conditional inputs can come from either image or text domain.}
    \vspace{1.5em}
    \label{fig:teaser}
}]

\maketitle

%%%%%%%%% ABSTRACT
\begin{abstract}
Hair editing is an interesting and challenging problem in computer vision and graphics. Many existing methods require well-drawn sketches or masks as conditional inputs for editing, however these interactions are neither straightforward nor efficient. In order to free users from the tedious interaction process, this paper proposes a new hair editing interaction mode, which enables manipulating hair attributes individually or jointly based on the texts or reference images provided by users. For this purpose, we encode the image and text conditions in a shared embedding space and propose a unified hair editing framework by leveraging the powerful image text representation capability of the Contrastive Language-Image Pre-Training (CLIP) model. With the carefully designed network structures and loss functions, our framework can perform high-quality hair editing in a disentangled manner. Extensive experiments demonstrate the superiority of our approach in terms of manipulation accuracy, visual realism of editing results, and irrelevant attribute preservation. Project repo is ~\url{https://github.com/wty-ustc/HairCLIP}.

\end{abstract}

%%%%%%%%% BODY TEXT
\section{Introduction}
\label{sec:intro}

Human hair, as the critical yet challenging component of the face, has long attracted the interest of researchers. In recent years, with the development of deep learning, many conditional GAN-based hair editing methods~\cite{tan2020michigan,xiao2021sketchhairsalon,Lee2020MaskGANTD} can produce satisfactory editing results. Most of these methods use well-drawn sketches~\cite{jo2019sc,tan2020michigan,xiao2021sketchhairsalon} or masks~\cite{Lee2020MaskGANTD,tan2020michigan} as the input of image-to-image translation networks to produce the manipulated results.

However, we think that these interaction types are not intuitive or user-friendly enough. For example, in order to edit the hairstyle of one image, users often need to spend several minutes to draw a good sketch, which greatly limits the large-scale, automated use of these methods. We therefore wonder ``\textit{Can we provide another more intuitive and convenient interaction way, just like human communication behaviors?}". And the language (or``text") naturally meets our requirements.

Benefiting from the development of cross-modal vision and language representations~\cite{tan2019lxmert, lu2019vilbert, su2019vl}, text-guided image manipulation has become possible. Recently, StyleCLIP~\cite{patashnik2021styleclip} has achieved amazing image manipulation results by leveraging the powerful image text representation capabilities of CLIP~\cite{Radford2021LearningTV}. %Contrastive Language-Image Pre-Training (CLIP) 
CLIP has an image encoder and a text encoder, by joint training on 400 million image text pairs, they can measure the semantic similarity between an input image and a text description. Based on this observation, StyleCLIP proposes to use them as the loss supervision to make the manipulated results match the text condition.

Although StyleCLIP inherently supports text description based hair editing, they are not exactly suitable for our task. It suffers from the following drawbacks: 1) For each specific hair editing description, it needs to train a separate mapper, which is not flexible in real applications; 2) The lack of tailored network structure and loss design makes the method poorly disentangled for hairstyle, hair color, and other unrelated attributes; 3) In practical applications, some hairstyles or colors are difficult to describe in text. At this time, users may prefer to use reference images, but StyleCLIP does not support reference image based hair editing.

To overcome the aforementioned limitations, we propose a hair editing framework that simultaneously supports different texts or reference images as the hairstyle/color conditions within one model. Generally, we follow StyleCLIP and utilize the StyleGAN~\cite{Karras2020AnalyzingAI} pre-trained on a large-scale face dataset as our generator, and then the key is to learn a mapper network to map the input conditions into corresponding latent code changes. But different from StyleCLIP, we explore the potential of CLIP to go beyond measuring image text similarity, along with some new designs: 1) \textit{Shared Condition Embedding}. To unify the text and image conditions into the same domain, we leverage the text encoder and image encoder of CLIP to extract their embedding as the conditions for the mapper network respectively. 2) \textit{Disentangled Information Injection}. We explicitly separate hairstyle and hair color information and feed them into different sub hair mappers corresponding to their semantic levels. This helps our method achieve disentangled hair editing; 3) \textit{Modulation Module}. We design a conditional modulation module to accomplish the direct control of input conditions on latent codes, which improves the manipulation ability of our method.

Since our goal is to achieve the hair editing based on the text or reference image condition while ensuring other irrelevant attributes unchanged, three types of losses are introduced: 1) Text manipulation loss is used to guarantee the similarity between the editing result and the given text description; 2) Image manipulation loss is used to guide hairstyle or hair color transfer from the reference image to the target image; 3) Attribute preservation loss is used to keep irrelevant attributes (e.g., identity and background) unchanged before and after editing.

Quantitative and qualitative comparisons and user study demonstrate the superiority of our method in terms of manipulation accuracy, manipulation fidelity, and irrelevant attribute preservation. And some example editing results are shown in \Fref{fig:teaser}. We also conduct extensive ablation analysis and well justify the designs of our network structure and loss function.

To summarize, our contributions are three-fold as below:
\begin{itemize}
	\item We push the frontiers of interactive hair editing, i.e., unifying text and reference image conditions within one framework. It supports a wide range of text and image conditions in one single model without the need of training many independent models, which has never been achieved before.
	
	\item In order to perform various hairstyle and hair color manipulation in a disentangled manner, we propose some new network structure designs and loss functions tailored for our task.

	\item Extensive experiments and analysis are conducted to show the better manipulation quality of our method and the necessity of each new design.
\end{itemize}

%-------------------------------------------------------------------------

\section{Related Work}

\noindent\textbf{Generative Adversarial Networks.} Since being proposed by Goodfellow \etal \cite{Goodfellow2014GenerativeAN}, GANs have made great progress in terms of loss functions \cite{Ansari2020ACF,Arjovsky2017WassersteinGA}, network structure design \cite{Schnfeld2020AUB,Gulrajani2017ImprovedTO,tan2021diverse}, and training strategies \cite{Tao2020AlleviationOG,Guo2020OnPC}. As a representative GAN in the field of image synthesis, StyleGAN \cite{Karras2019ASG,Karras2020AnalyzingAI} can synthesize very high-fidelity human faces with realistic facial details and hair. As the typical unconditional GANs, StyleGAN itself is difficult to achieve controllable image synthesis effects. But fortunately, its latent space demonstrates promising disentanglement properties \cite{Collins2020EditingIS, Shen2020InterpretingTL, Goetschalckx2019GANalyzeTV, Jahanian2020OnT}, and many works utilize StyleGAN to perform image manipulation tasks \cite{wang2021cross,Nitzan2020FaceID, Wei2021ASB,Alaluf2021OnlyAM, patashnik2021styleclip}. In this paper, we convert the unconditional StyleGAN into our conditional hair editing network with the help of CLIP's powerful image text representation capability. Moreover, we unify the text and reference image condition in one framework and achieve disentangled editing effects.

\noindent\textbf{Image-based Hair Manipulation.} As an important part of the human face, hair has attracted many works dedicated to hair modeling \cite{Hu2014RobustHC, Chai2016AutoHairFA, Chai2015HighqualityHM} and synthesis \cite{Wei2018RealTimeHR, Jo2019SCFEGANFE, Lee2020MaskGANTD, Zhu2020SEANIS}. Some works \cite{Lee2020MaskGANTD, Zhu2020SEANIS} use mask which explicitly decouples facial attributes including hair as the conditional input for image-to-image translation networks to accomplish hair manipulation. There are also several works \cite{tan2020michigan, xiao2021sketchhairsalon} that use sketches as input to depict the structure and shape of the desired hairstyle. However, such interactions are still relatively costly for users. To enable easier interaction, MichiGAN \cite{tan2020michigan} supports hair transfer by extracting the orientation map of one hairstyle reference image as well as the appearance from another hair color reference image. However, MichiGAN is easy to fail for arbitrary shape changes during hair transfer. Recently, LOHO~\cite{Saha2021LOHOLO} performs a two-stage optimization in the $ \mathcal{W+} $ space and noise space of StyleGANv2~\cite{Karras2020AnalyzingAI} to complete the hair transfer for a given reference image. However, the area optimized by this method is limited to the foreground, which requires blending the reconstructed foreground with the original background and often brings obvious artifacts. Besides, it is very time-consuming, e.g., several minutes to optimize an image.

\noindent\textbf{Text-based Hair Manipulation.} Along with the booming development of cross-modal visual and language representations~\cite{tan2019lxmert, lu2019vilbert, su2019vl}, especially the powerful CLIP~\cite{Radford2021LearningTV}, many recent efforts ~\cite{chen2018language, jiang2021talk, xia2021tedigan, patashnik2021styleclip} start to study text based manipulation. However, there is no existing method specifically tailored for hair editing. Among these works, the most relevant to our work are StyleCLIP~\cite{patashnik2021styleclip} and TediGAN~\cite{xia2021tedigan}. But StyleCLIP needs to train  a separate mapper network for each specific hair editing description, which is not
flexible for real applications. For TediGAN, it proposes two approaches: TediGAN-A encodes text and image separately into the latent space of StyleGAN and completes manipulation with style-mixing, which is less decoupled and difficult to complete hair editing; TediGAN-B accomplishes the manipulation with optimization using CLIP to provide text-image similarity loss, but the lack of knowledge learned from a large dataset makes the process unstable and time-consuming. 

Different from existing works, this paper presents the first unified framework that enables the text and image conditions simultaneously. This provides a more intuitive and convenient interaction mode, and enables diverse text and image conditions within one single model. Besides, benefiting from the new designs tailored for this task, our method also shows much better hair manipulation quality.

\section{Proposed Method}
\begin{figure*}[t]
	\centering
	\includegraphics[width=\textwidth]{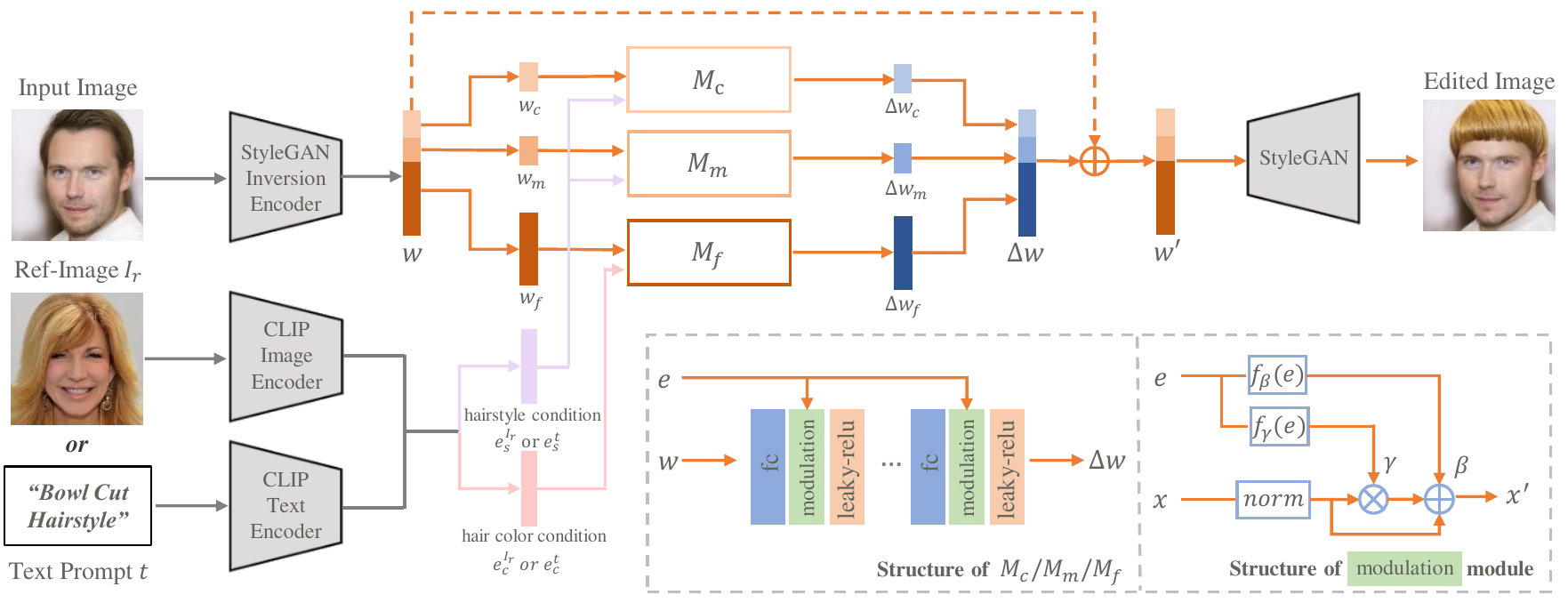} 
	\caption{The overview of our framework, here we show an example with hairstyle description text and hair color reference image as conditional inputs. Our framework supports to accomplish the corresponding hair editing according to the given reference images and texts, where images, texts are encoded by CLIP's image encoder, text encoder to 512-dimensional vectors as conditional inputs for the hair mapper, respectively. Only three sub hair mappers are trainable, where $ M_{c} $ and $ M_{m} $ take the hairstyle conditional input $ e_{s} $ and $ M_{f} $ takes the hair color conditional input $ e_{c} $.} 
	\label{fig:modelfigure}
	\vspace{-1em}
\end{figure*}

\subsection{Overview}
Imagine we are in a barbershop and if someone wants to design his hair, the common interaction would be to name the desired hairstyle or provide the hairstylist with a corresponding picture. Inspired by this, we think empowering the AI algorithms to enable such an intuitive and efficient interaction mode is really needed. Thanks to the great image synthesis quality of StyleGAN ~\cite{Karras2019ASG,Karras2020AnalyzingAI} and the excellent image/text representation ability of CLIP ~\cite{Radford2021LearningTV}, we are finally able to design such a unified hair editing framework to achieve this goal. Before diving into the framework details, we briefly introduce StyleGAN and CLIP respectively.

\noindent\textbf{StyleGAN} \cite{Karras2019ASG,Karras2020AnalyzingAI}  can synthesize high-resolution, high-fidelity realistic images with a progressive upsample network from noises. Its synthesis process involves multiple latent spaces. $ \mathcal{Z} \in \mathbb{R}^{512} $ is the original noise space of StyleGAN. A randomly sampled noise vector $ z \in \mathcal{Z}$ is transformed to the $ \mathcal{W} \in \mathbb{R}^{512} $ latent space after $ 8 $ fully connected layers. Several studies \cite{Collins2020EditingIS, Shen2020InterpretingTL, Goetschalckx2019GANalyzeTV, Jahanian2020OnT} have demonstrated that StyleGAN spontaneously learns to encode rich semantics within its $ \mathcal{W} $ space during training, and thus $ \mathcal{W} $ exhibits good semantic decoupling properties. In addition, some recent StyleGAN inversion works \cite{Abdal2019Image2StyleGANHT, richardson2021encoding, Wei2021ASB} extend $\mathcal{W} $ space to $ \mathcal{W+} $ space for better reconstruction. For a StyleGAN with 18 layers, it is defined by the cascade of $ 18 $ different $ 512 $-dimensional vectors $ [w_1, ..., w_{18}], w_i \in \mathcal{W}$. 

\noindent\textbf{CLIP}~\cite{Radford2021LearningTV} is a multi-modality model pretrained from $ 400 $ million image-text pairs collected from the Internet.  It consists of one image encoder and one text encoder that will encode the image and text into the $512$-dimensional embedding vector, respectively. It adopts the typical contrastive learning framework, which minimizes the cosine distance between the encoded vectors of the correct image text pairs and maximizes the cosine distance of the incorrect pairs. Benefiting from large-scale pretraining, CLIP can well measure the semantic similarity between an image and a text, via learning one shared image-text embedding space.

\subsection{HairCLIP}
Inspired by the pioneering work StyleCLIP\cite{patashnik2021styleclip}, we utilize the powerful synthesis ability of the pretrained StyleGAN, and aim to learn an extra mapper network to achieve the hair editing function. More specifically, given the real image to edit, we first use the StyleGAN inversion method ``e4e" \cite{Tov2021DesigningAE} to get its latent code $w$ in the $\mathcal{W+}$ space, then use the mapper network to predict the latent code change $\Delta w$ based on $w$ and editing conditions (including hairstyle condition $e_s$ and hair color condition $e_c$). Finally, the modified latent code $w'=w+\Delta w$ will be fed back into the pretrained StyleGAN to get the target editing result. The overall pipeline is illustrated in \Fref{fig:modelfigure}, and each component will be elaborated below.

\noindent\textbf{Shared Condition Embedding.} To unify the conditions from the text and image domains under one framework, we naturally choose to represent them by embedding them in the joint latent space of CLIP. For the user-supplied text hairstyle prompt and text hair color prompt, we use CLIP's text encoder to encode them into $512$-dimensional conditional embedding, which are denoted as $e_{s}^{t}$ and $e_{c}^{t}$ respectively. Similarly, the hairstyle reference image and hair color reference image are encoded by the image encoder of CLIP and denoted as $e_{s}^{I_{r}}$ and $e_{c}^{I_{r}}$ respectively. Because CLIP is well trained on large-scale image-text pairs, $e_{s}^{t},e_{c}^{t},e_{s}^{I_{r}},e_{c}^{I_{r}}$ all reside in the shared latent space, thus can be fed into one mapper network and flexibly switched.

\noindent\textbf{Disentangled Information Injection.} As demonstrated in many works~\cite{xia2021tedigan,Karras2019ASG}, different layers of StyleGAN correspond to different semantic levels of information in the generated images, with the more preceding layers corresponding to higher semantic levels of information. Following the StyleCLIP \cite{patashnik2021styleclip}, we adopt three sub hair mappers $ M_{c} $, $ M_{m}, M_{f}$ with the same network structure, which are responsible for predicting $ \Delta w $ of hair editing corresponding to different parts (coarse, medium and fine) of the latent code $ w=(w_{c}, w_{m}, w_{f}) $. More specifically, $ w_{c}, w_{m}, w_{f}$ correspond to the high semantic level, the middle semantic level, and the low semantic level respectively.

Noticing this semantic layering phenomenon in StyleGAN, we propose disentangled information injection, which aims to improve the decoupling ability of the network for hairstyle and hair color editing. In detail, we use the embedding of hairstyle information $ e_{s} \in \{e_{s}^{t}, e_{s}^{I_{r}}\} $ from CLIP as the conditional input for $ M_{c} $ and $ M_{m} $, and the embedding of hair color information $ e_{c} \in \{e_{c}^{t}, e_{c}^{I_{r}}\} $ from CLIP as the conditional input for $ M_{f} $. This is based on the empirical observation that
 hairstyle often corresponds to middle and high level semantic information in StyleGAN while hair color corresponds to low level semantic information.  Therefore, the hair mapper $ M $ can be formulated as:

\begin{equation}
M(w,e_{s},e_{c})=(M_{c}(w_{c},e_{s}),M_{m}(w_{m},e_{s}),M_{f}(w_{f},e_{c})).
\end{equation}

\noindent\textbf{Modulation Module.} As shown in Figure \ref{fig:modelfigure}, each sub hair mapper network follows a simple design and consists of five blocks, and each block consists of one fully connected (fc) layer, one newly designed modulation module, and one non-linear activation layer (leakly relu). Rather than simply concatenating the condition embedding with the input latent code, the modulation module uses the condition embedding $e$ to modulate the intermediate output $x$ of the preceding fc layer. Mathematically, it follows the below formulation:
\begin{equation}
x' =(1+f_{\gamma}(e))\dfrac{x-\mu_x}{\sigma_x}+f_{\beta}(e),
\end{equation}
where $\mu_x$ and $\sigma_x$ denote the mean and standard deviation of $x$ respectively. And $ f_{\gamma} $ and $ f_{\beta}$ are implemented with simple fully connected networks (two fc layers with one intermediate layernorm and leaky relu layer). This design is motivated by recent conditional image translation works \cite{park2019semantic,tan2021efficient,huang2017arbitrary}. During testing, if no conditional input is provided for hairstyle or hair color, then all modulation modules in the corresponding sub hair mapper will be implemented as identity functions, and we denote this case as $ e_{s}=0 $ or $ e_{c}=0 $. In this way, we flexibly support users to edit only hairstyle, only hair color, or both hairstyle and hair color.

\subsection{Loss Functions}
Our goal is to manipulate the hair in a decoupled manner based on the conditional input, while requiring other irrelevant attributes (e.g., background, identity) well preserved. Therefore, we specifically design three types of loss functions to train the mapper networks: text manipulation loss, image manipulation loss, and attribute preservation loss.

\noindent\textbf{Text Manipulation Loss.} In order to perform the corresponding hair manipulation based on the text prompt of the hairstyle or color, we design the text manipulation loss $ \mathcal{L}_{t} $ with the help of CLIP as follows:
\begin{equation}
\mathcal{L}_{t}=\mathcal{L}_{st}^{clip} + \mathcal{L}_{ct}^{clip}.
\end{equation}
For the hairstyle text manipulation loss, we measure the cosine distance between the manipulated image and the given text in the CLIP's latent space:
\begin{equation}
\mathcal{L}_{st}^{clip}=1-cos(E_{i}(G(w+M(w,e_{s}^{t},e_{c}))), e_{s}^{t}),
\end{equation}
where $ cos(\cdot)$ means cosine similarity, $ E_{i} $ represents the image encoder of CLIP, $ G $ represents the pretrained StyleGAN generator, $ e_{s}^{t}=E_{t}(st) $ denotes the embedding of a given hairstyle description text $ st $ which is encoded by the text encoder $ E_{t} $ of CLIP, and $ e_{c} \in \{e_{c}^{t}, e_{c}^{I_{r}}, 0\} $. Similarly, color text manipulation loss is defined as follows:
\begin{equation}
\mathcal{L}_{ct}^{clip}=1-cos(E_{i}(G(w+M(w,e_{s},e_{c}^{t}))), e_{c}^{t}),
\end{equation}
where $ e_{c}^{t} $ denotes the embedding of a given color description text which is encoded by the text encoder of CLIP, and $ e_{s} \in \{e_{s}^{t}, e_{s}^{I_{r}}, 0\} $.

\noindent\textbf{Image Manipulation Loss.} Given a reference image, we want the manipulated image to possess the same hairstyle as that of the reference image. However characterizing the similarity between two hairstyles is a challenging task. Exploiting the powerful potential of CLIP again, we encode them separately using CLIP's image encoder to measure their similarity in CLIP's latent space:
\begin{equation}
\mathcal{L}_{si}=1-cos(E_{i}(\mathbf{x}_{M}\ast P_{h}(\mathbf{x}_{M})), E_{i}(\mathbf{x}\ast P_{h}(\mathbf{x}))),
\end{equation}
where the manipulated image $ \mathbf{x}_{M}=G(w+M(w,e_{s}^{I_{r}},e_{c})) $, $ e_{s}^{I_{r}}=E_{i}(\mathbf{x}\ast P_{h}(\mathbf{x})) $, $ e_{c} \in \{e_{c}^{t}, e_{c}^{I_{r}}, 0\} $, $ P $ denotes the pre-trained facial parsing network~\cite{FaceParsing},  $ P_{h}(\mathbf{x}_{M}) $ represents the mask of the hair region of $ \mathbf{x}_{M} $, and $ \mathbf{x} $ means the given reference image. Thanks to this supervision we propose, our method can yield plausible editing results for cases where the reference image and the input image are seriously misaligned, which is currently unavailable for other hairstyle transfer methods. Also, for reference image based hair color manipulation, we calculate the average color difference in the hair area between reference image and manipulated image as the loss:
\begin{equation}
\mathcal{L}_{ci}=||avg(\mathbf{x}_{M}\ast P_{h}(\mathbf{x}_{M}))-avg(\mathbf{x}\ast P_{h}(\mathbf{x}))||_{1},
\end{equation}
where $ \mathbf{x}_{M}=G(w+M(w,e_{s},e_{c}^{I_{r}})) $, $ e_{c}^{I_{r}}=E_{i}(\mathbf{x}\ast P_{h}(\mathbf{x})) $, and $ e_{s} \in \{e_{s}^{t}, e_{s}^{I_{r}}, 0\} $.
In summary, the image manipulation loss $ \mathcal{L}_{i} $ is defined as:
\begin{equation}
\mathcal{L}_{i}=\lambda_{si}\mathcal{L}_{si} + \lambda_{ci}\mathcal{L}_{ci},
\end{equation}
where $ \lambda_{si} $, $ \lambda_{ci} $ are set to $ 5 $, $ 0.02 $ respectively by default.

\noindent\textbf{Attribute Preservation Loss.} To ensure identity consistency before and after hair editing, the identity loss is applied as follows:
\begin{equation}
\mathcal{L}_{id}=1-cos(R(G(w+M(w,e_{s},e_{c}))), R(G(w))),
\end{equation}
where $ e_{s} \in \{e_{s}^{t}, e_{s}^{I_{r}}, 0\} $, $ e_{c} \in \{e_{c}^{t}, e_{c}^{I_{r}}, 0\} $, $ R $ is a pretrained ArcFace~\cite{Deng2019ArcFaceAA} network for face recognition and $ G(w) $ denotes the reconstructed real image. In addition, we designed $ \mathcal{L}_{s\_mc} $ in the same way as $ \mathcal{L}_{ci} $ in order to maintain the hair color when only manipulating the hairstyle:
\begin{equation}
\mathcal{L}_{s\_mc}=||avg(\mathbf{x}_{M}\ast P_{h}(\mathbf{x}_{M}))-avg(\mathbf{x}_{w}\ast P_{h}(\mathbf{x}_{w}))||_{1},
\end{equation}
where $ \mathbf{x}_{M}=G(w+M(w,e_{s},e_{c})) $, $ e_{s} \in \{e_{s}^{t}, e_{s}^{I_{r}}\} $, $ e_{c}=0 $, and $ \mathbf{x}_{w}=G(w) $. Empirically, we find the hairstyle can be well preserved when only changing the color, so we do not add corresponding preservation loss.

Moreover, we introduced background loss with the help of facial parsing network~\cite{FaceParsing} :
\begin{equation}
\mathcal{L}_{bg}=||(\mathbf{x}_{M}-\mathbf{x}_{w})\ast(P_{nh}(\mathbf{x}_{M})\cap P_{nh}(\mathbf{x}_{w}))||_{2},
\end{equation}
where $ \mathbf{x}_{M}=G(w+M(w,e_{s},e_{c})) $, $ P_{nh}(\mathbf{x}_{M}) $ represents the mask of the non-hair region of $ \mathbf{x}_{M} $. In this way, we largely ensure that the non-relevant attribute regions remain unchanged. For the same purpose, the $ L_{2} $ norm of the manipulation step in the latent space is utilized:
\begin{equation}
\mathcal{L}_{norm}=||M(w,e_{s},e_{c})||_{2}.
\end{equation}
The overall attribute preservation loss $ \mathcal{L}_{ap} $ is defined as:
\begin{equation}
\mathcal{L}_{ap}=\lambda_{id}\mathcal{L}_{id} + \lambda_{s\_mc}\mathcal{L}_{s\_mc}+\lambda_{bg}\mathcal{L}_{bg}+\lambda_{norm}\mathcal{L}_{norm},
\end{equation}
where $ \lambda_{id} $, $ \lambda_{s\_mc} $, $ \lambda_{bg} $, $ \lambda_{norm} $ are set to $ 0.3 $, $ 0.02 $, $ 1 $, $ 0.8 $ respectively by default.

Finally, the overall loss function is defined as:
\begin{equation}
\mathcal{L}=\lambda_{t}\mathcal{L}_{t} + \lambda_{i}\mathcal{L}_{i}+ \lambda_{ap}\mathcal{L}_{ap},
\end{equation}
where $ \lambda_{t} $, $ \lambda_{i} $, $ \lambda_{ap} $ are set to $ 2 $, $ 1 $, $ 1 $ respectively by default.

\section{Experiments}

\begin{figure*}[tb]
	\begin{center}
		\setlength{\tabcolsep}{0.5pt}
		
		\begin{tabular}{m{0.3cm}<{\centering}m{2.05cm}<{\centering}m{2.05cm}<{\centering}m{2.05cm}<{\centering}m{2.05cm}<{\centering}m{0.3cm}<{\centering}m{2.05cm}<{\centering}m{2.05cm}<{\centering}m{2.05cm}<{\centering}m{2.05cm}<{\centering}}
			& \small{Input Image} & \small{Ours} & \small{StyleCLIP~\cite{patashnik2021styleclip}} & \small{TediGAN~\cite{xia2021tedigan}} &  & \small{Input Image} & \small{Ours} & \small{StyleCLIP~\cite{patashnik2021styleclip}} & \small{TediGAN~\cite{xia2021tedigan}}
			\\
			
			\raisebox{0.7cm}{\rotatebox[origin=c]{90}{\footnotesize{{afro hairstyle}}}}
			&\includegraphics[width=2cm]{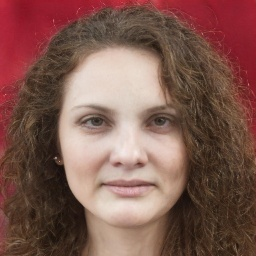}
			&\includegraphics[width=2cm]{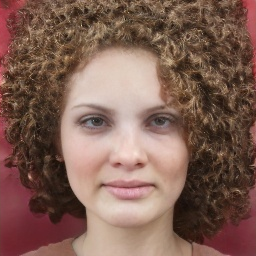}
			&\includegraphics[width=2cm]{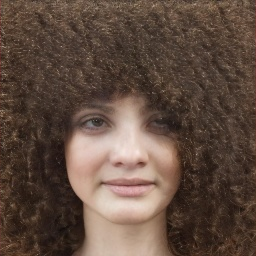}
			&\includegraphics[width=2cm]{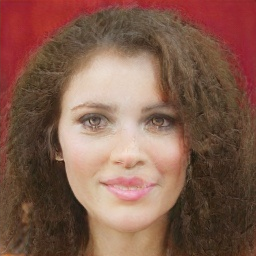}
			&\raisebox{0.5cm}{\rotatebox[origin=c]{90}{\footnotesize{{green hair}}}}
			&\includegraphics[width=2cm]{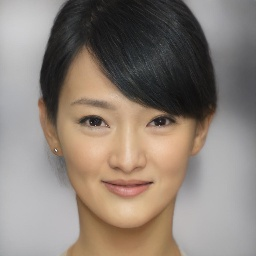}
			&\includegraphics[width=2cm]{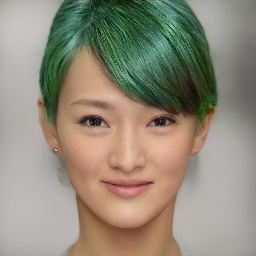}
			&\includegraphics[width=2cm]{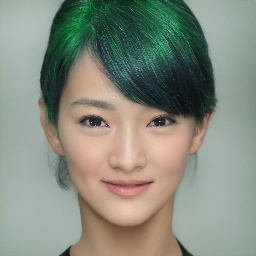}
			&\includegraphics[width=2cm]{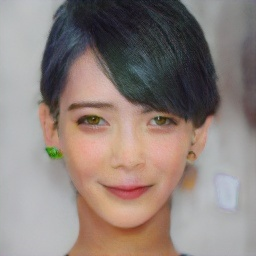}			
			\\

			\raisebox{0.83cm}{\rotatebox[origin=c]{90}{\footnotesize{{bobcut hairstyle}}}}
			&\includegraphics[width=2cm]{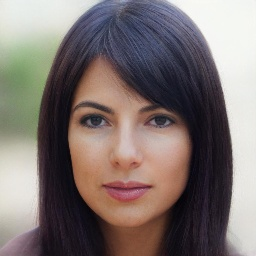}
			&\includegraphics[width=2cm]{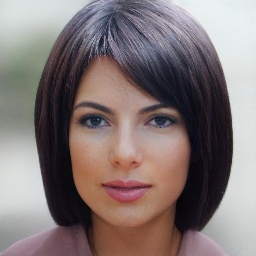}
			&\includegraphics[width=2cm]{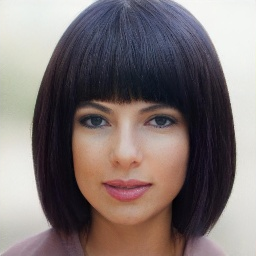}
			&\includegraphics[width=2cm]{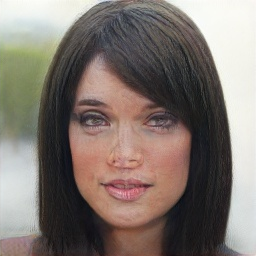}
			&\raisebox{0.5cm}{\rotatebox[origin=c]{90}{\footnotesize{{blond hair}}}}
			&\includegraphics[width=2cm]{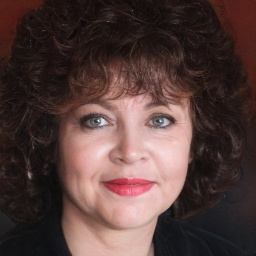}
			&\includegraphics[width=2cm]{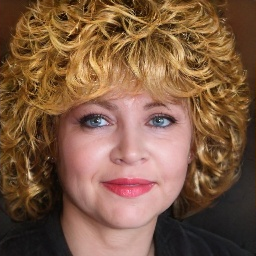}
			&\includegraphics[width=2cm]{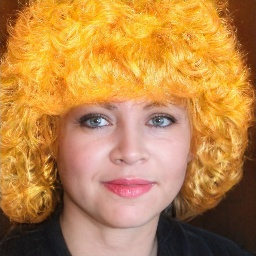}
			&\includegraphics[width=2cm]{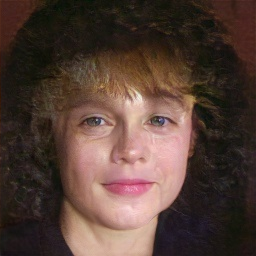}			
			\\
			
			\raisebox{0.87cm}{\rotatebox[origin=c]{90}{\footnotesize{{bowlcut hairstyle}}}}
			&\includegraphics[width=2cm]{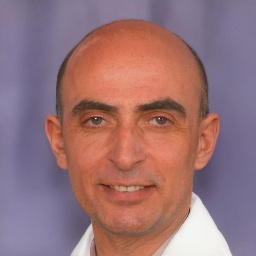}
			&\includegraphics[width=2cm]{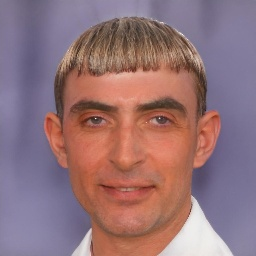}
			&\includegraphics[width=2cm]{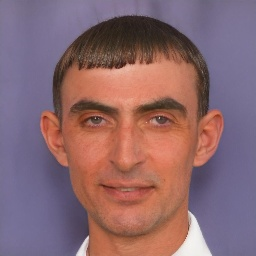}
			&\includegraphics[width=2cm]{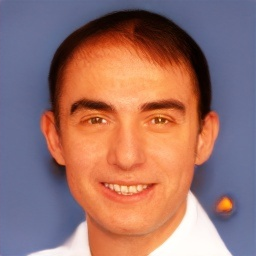}
			&\raisebox{0.5cm}{\rotatebox[origin=c]{90}{\footnotesize{{braid brown}}}}
			&\includegraphics[width=2cm]{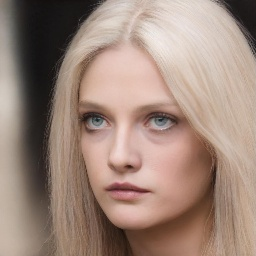}
			&\includegraphics[width=2cm]{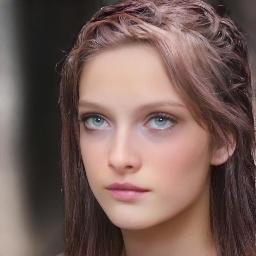}
			&\includegraphics[width=2cm]{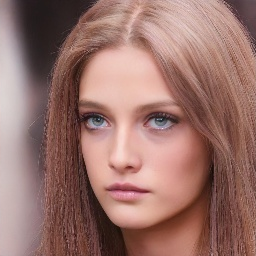}
			&\includegraphics[width=2cm]{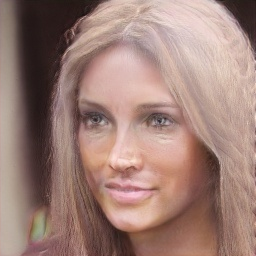}
			\\
			
			\raisebox{0.93cm}{\rotatebox[origin=c]{90}{\footnotesize{{mohawk hairstyle}}}}
			&\includegraphics[width=2cm]{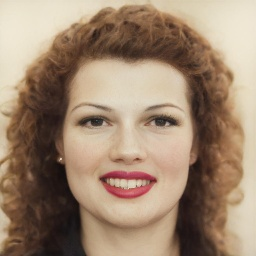}
			&\includegraphics[width=2cm]{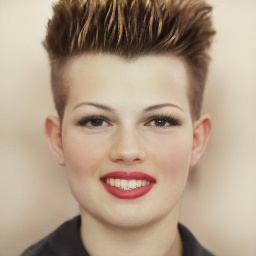}
			&\includegraphics[width=2cm]{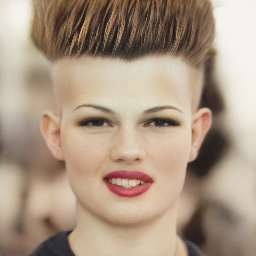}
			&\includegraphics[width=2cm]{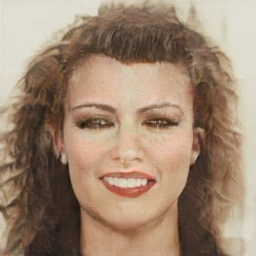}
			&\raisebox{0.75cm}{\rotatebox[origin=c]{90}{\footnotesize{{crewcut yellow}}}}
			&\includegraphics[width=2cm]{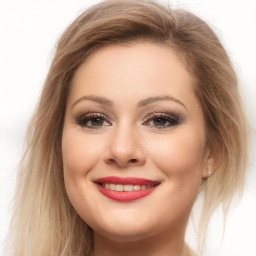}
			&\includegraphics[width=2cm]{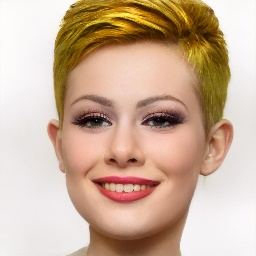}
			&\includegraphics[width=2cm]{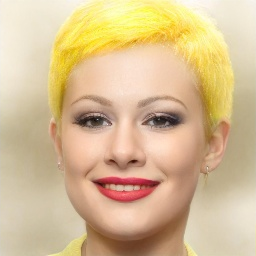}
			&\includegraphics[width=2cm]{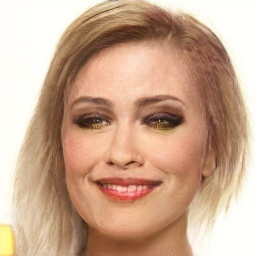}
			\\
			
			\raisebox{0.51cm}{\rotatebox[origin=c]{90}{\footnotesize{{purple hair}}}}
			&\includegraphics[width=2cm]{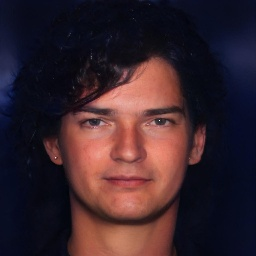}
			&\includegraphics[width=2cm]{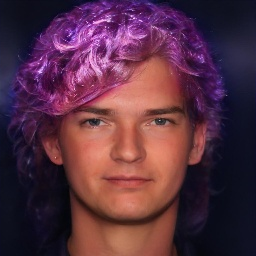}
			&\includegraphics[width=2cm]{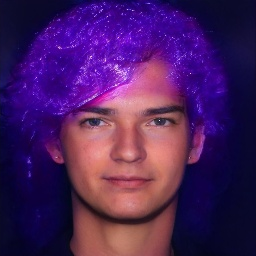}
			&\includegraphics[width=2cm]{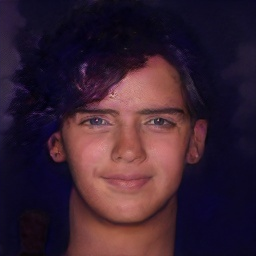}
			&\raisebox{0.51cm}{\rotatebox[origin=c]{90}{\footnotesize{{perm gray}}}}
			&\includegraphics[width=2cm]{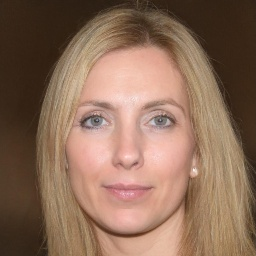}
			&\includegraphics[width=2cm]{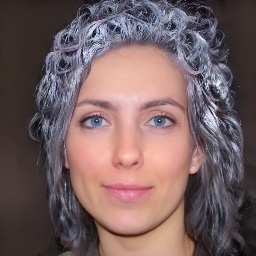}
			&\includegraphics[width=2cm]{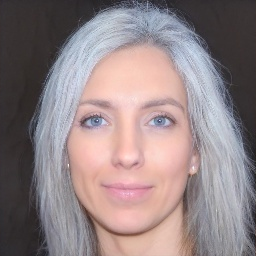}
			&\includegraphics[width=2cm]{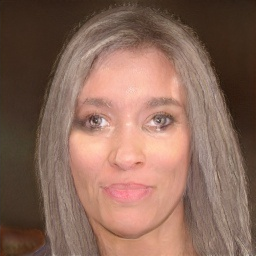}
			\\
			
		\end{tabular}
	\end{center}
	\vspace{-0.8em}
	\caption{Visual comparison with StyleCLIP~\cite{patashnik2021styleclip} and TediGAN~\cite{xia2021tedigan}. The corresponding simplified text descriptions (editing hairstyle, hair color, or both of them) are listed on the leftmost side of each row, and all input images are the inversions of the real images. Our approach demonstrates better visual photorealism and irrelevant attributes preservation ability while completing the specified hair editing.} 
	\label{fig:textcomparefig}
	\vspace{-0.5em}
\end{figure*}

\noindent\textbf{Implementation Details.} We train and evaluate our hair mapper on the CelebA-HQ dataset \cite{karras2018progressive}. Since we use e4e \cite{Tov2021DesigningAE} as our inversion encoder, we follow its division of the training set and test set. The StyleGAN2 \cite{Karras2020AnalyzingAI} pre-trained on the FFHQ dataset \cite{Karras2019ASG} is used as our generator. For the text input, we collected $ 44 $ hairstyle text descriptions and $ 12 $ hair color text descriptions; The CelebA-HQ dataset is used to provide reference images of hairstyles or hair colors, and we also generated several edited images using our text-guided hair editing method to augment the diversity of the reference image set. During training, the hair mapper is randomly tasked to edit only the hairstyle or only the hair color or both hairstyle and hair color depending on the provided conditional input. The conditioned input is randomly set as text or reference image. Regarding the training strategy, the base learning rate is $0.0005$ with batch size of $ 1 $. The number of training iterations is $ 500,000 $, and the Adam \cite{kingma2015adam} optimizer is used, with $ \beta_{1} $ and $ \beta_{2} $ set to $ 0.9 $ and $ 0.999 $, respectively. For all compared methods, we use the official training codes or pre-trained models.

To quantitatively evaluate irrelevant attributes preservation, four metrics are used: IDS denotes identity similarity before and after editing calculated by Curricularface \cite{Huang2020CurricularFaceAC}. PSNR and SSIM are calculated in the region of intersection of non-hair regions before and after editing. ACD represents the average color difference of the hair region.

\subsection{Quantitative and Qualitative Comparison}

\noindent\textbf{Comparison to Text-Driven Image Manipulation Methods.} We compare our approach with current state-of-the-art text-driven image manipulation methods TediGAN~\cite{xia2021tedigan} and StyleCLIP~\cite{patashnik2021styleclip} on ten text descriptions. The optimization iteration number of TediGAN is set to $ 200 $ according to their official recommendations. The visual comparison is shown in Figure \ref{fig:textcomparefig}. TediGAN fails in all hairstyle editing related tasks, only the hair color editing is barely successful but the results are still unsatisfactory. This phenomenon is consistent with the findings given in the StyleCLIP paper: the optimization method using CLIP similarity loss is very unstable due to the lack of knowledge learned from a large dataset.

StyleCLIP trains a separate mapper for each description and thus demonstrates stronger manipulation ability on the task of editing only the hairstyle, but excessive manipulation ability instead affects the image realism (see afro hairstyle). Thanks to our \textit{shared condition embedding}, our method finds a balance between the degree of manipulation and realism by fully learning over many hair editing description inputs. On the task of editing both hairstyle and hair color, our method exhibits better manipulation ability. This is due to proposed \textit{disentangled information injection} and \textit{modulation module}, whereas StyleCLIP leaves this information in one description making it poorly decoupled and difficult to perform hairstyle and hair color editing tasks at the same time. In addition, benefiting from attribute preservation loss, our method exhibits better retention of irrelevant attributes (see mohawk hairstyle, purple hair). 

In Table \ref{tab:textcomparetable}, we give the average quantitative comparison results in terms of irrelevant attributes preservation on these ten text descriptions. And the quantitative results lead to the same conclusions as the visual comparison. We do not compare the FID~\cite{Heusel2017GANsTB} used in TediGAN here since it can not reflect the manipulation capability. More quantitative results and analysis in terms of the FID metric 
are given in the supplementary material.

\begin{table}[t]
	\centering
	\setlength{\tabcolsep}{1em}{
		\begin{tabular}{lccc}
			\hline
			Methods & IDS & PSNR & SSIM \\
			\hline
			Ours & \textbf{0.83} & \textbf{27.8} & \textbf{0.92}  \\
			StyleCLIP~\cite{patashnik2021styleclip} & 0.79 & 23.2 & 0.87  \\
			TediGAN~\cite{xia2021tedigan} & 0.17 & 24.1 & 0.79  \\
			\hline
		\end{tabular}
	}
	\caption{Quantitative comparison regarding the preservation of irrelevant attributes. Our approach exhibits the best irrelevant attributes preservation ability.}	
	\label{tab:textcomparetable}

\end{table}

\begin{figure}[t]
	\begin{center}
		\setlength{\tabcolsep}{0.5pt}
		\begin{tabular}{cccccc}
			\small{Input} & \small{HRI} & \small{CRI} & \small{Ours} & \small{LOHO} & \small{MichiGAN}
			\\
			\includegraphics[width=1.34cm]{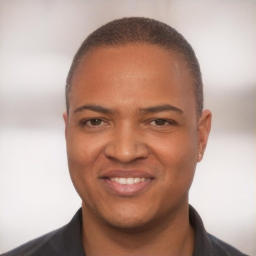}
			&\includegraphics[width=1.34cm]{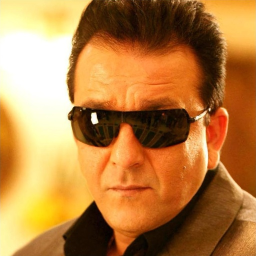}
			&\includegraphics[width=1.34cm]{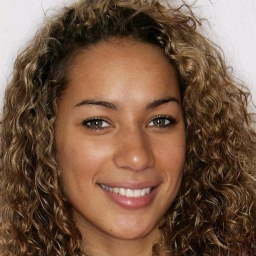}
			&\includegraphics[width=1.34cm]{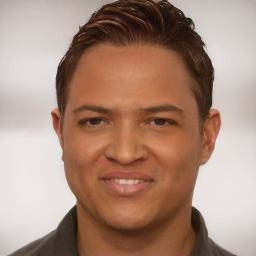}
			&\includegraphics[width=1.34cm]{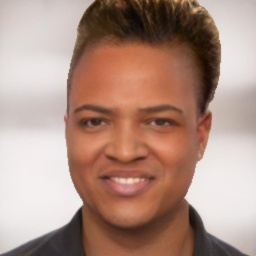}
			&\includegraphics[width=1.34cm]{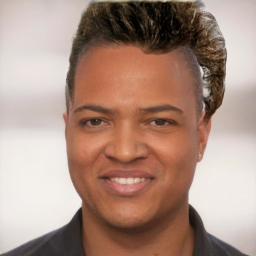}
			\\
			
			\includegraphics[width=1.34cm]{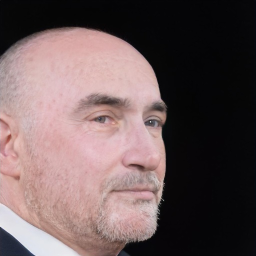}
			&\includegraphics[width=1.34cm]{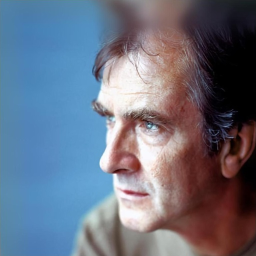}
			&\includegraphics[width=1.34cm]{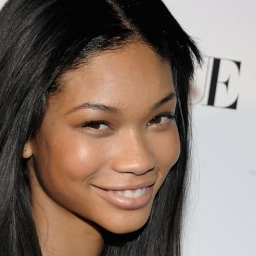}
			&\includegraphics[width=1.34cm]{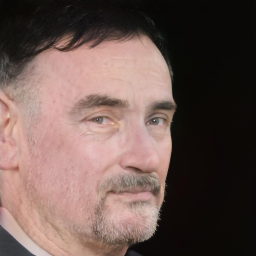}
			&\includegraphics[width=1.34cm]{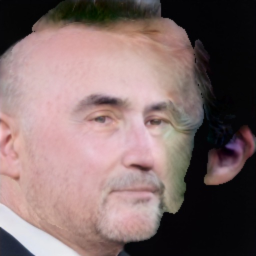}
			&\includegraphics[width=1.34cm]{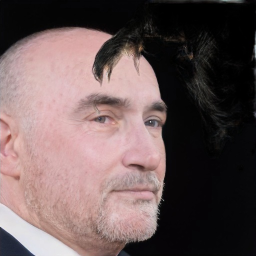}
			\\
			
			\includegraphics[width=1.34cm]{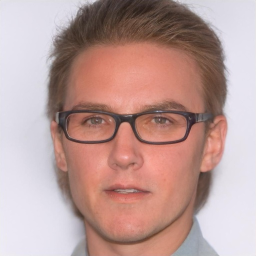}
			&\includegraphics[width=1.34cm]{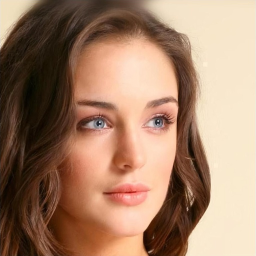}
			&\includegraphics[width=1.34cm]{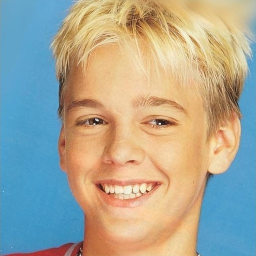}
			&\includegraphics[width=1.34cm]{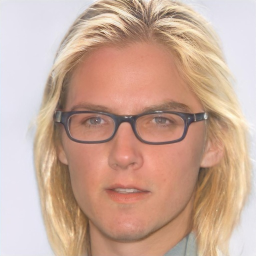}
			&\includegraphics[width=1.34cm]{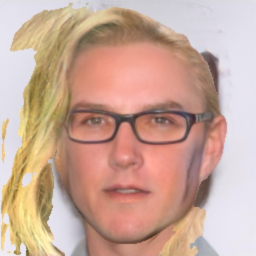}
			&\includegraphics[width=1.34cm]{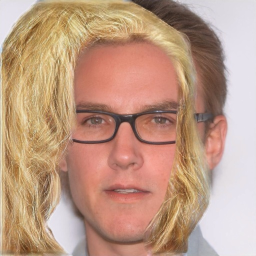}
			\\
		\end{tabular}
	\end{center}

	\caption{Comparison of our approach with LOHO~\cite{Saha2021LOHOLO} and MichiGAN~\cite{tan2020michigan} on hair transfer. HRI means hairstyle reference image and CRI means hair color reference image.} 
	\label{fig:transfer-compare}

\end{figure}

\noindent\textbf{Comparison to Hair Transfer Methods.} Given a hairstyle reference image and a hair color reference image, the purpose of hair transfer is to transfer their corresponding hairstyle and hair color attributes to the input image. We compare our method with the current state-of-the-art  LOHO~\cite{Saha2021LOHOLO} and MichiGAN~\cite{tan2020michigan} in Figure \ref{fig:transfer-compare}. Both of these methods perform hairstyle transfer by direct replication in the spatial domain to generate more accurate details of the hair structure, although suffer from obvious artifacts in the boundary areas in some cases (see the results in the first row). However, as shown in the last two rows, they are sensitive to the pose of hairstyle reference images and cannot complete plausible hairstyle transfer when the hairstyle and pose are not well aligned between the hairstyle reference image and the input image. Unlike these two approaches, we transform the measure space of similarity into the latent space of CLIP during training and use the embedding of the hair region of the reference image from CLIP as the conditional input. As a result, our method provides a solution for the unaligned hairstyle transfer and shows its superiority compared to other existing methods.

\noindent\textbf{User Study.} To further evaluate the manipulation ability and the visual realism of the edited results of different methods in two types of hair editing tasks, we recruited $ 20 $ participants for our user study. For the text-driven image manipulation methods, we provided $ 20 $ groups of results from three methods at a time, which were randomly selected from two of each of ten hair editing descriptions. For the hair transfer methods, participants were also provided with $ 20 $ groups of results, half of which were aligned hairstyle transfer cases and the other half were non-aligned. Participants were asked to rank three methods for each task with respect to manipulation accuracy and visual realism, where $ 1 $ represents the best and $ 3 $ represents the worst. The average ranking values are listed in Table \ref{tab:userstudy}, where our method outperforms the competitive approaches in both metrics.

\begin{table}[t]
	\centering
	\small
	\setlength{\tabcolsep}{1.8pt}{
		\begin{tabular}{l|ccc|ccc}
			\hline
			\multicolumn{1}{ c }{{}}	& \multicolumn{3}{ c }{{\small{Text-Driven Methods}}} & \multicolumn{3}{ c }{{\small{Hair Transfer Methods}}} \\
			\hline
			\small{Metrics} & \small{Ours} & \small{StyleCLIP} & \small{TediGAN} & \small{Ours} & \small{LOHO} & \small{MichiGAN}\\
			\hline
			Acc.  & \textbf{\small{1.39}} & \small{1.66} & \small{2.95} & \textbf{\small{1.79}} & \small{2.26} & \small{1.95} \\
			Real.  & \textbf{\small{1.42}} & \small{1.63} & \small{2.95} & \textbf{\small{1.09}} & \small{2.48} & \small{2.43} \\
			\hline
		\end{tabular}
	}
	\caption{User study on text-driven image manipulation methods and hair transfer methods. Acc. denotes the manipulation accuracy for given conditional inputs and Real. denotes the visual realism of the manipulated image. The numbers in the table are average rankings, the lower the better.}	
	\label{tab:userstudy}
\end{table}

\subsection{Ablation Analysis}
To verify the effectiveness of our proposed network structure and loss functions, we alternately ablate one of these key components to retrain variants of our method, by keeping all but the selected component unchanged.

\noindent\textbf{Importance of Attribute Preservation Loss.} To verify the role of each component in the attribute preservation loss, we randomly selected $ 4,400 $ images for qualitative and quantitative ablation studies across the task of editing only hairstyles. Consistent conclusions can be drawn from Table \ref{tab:quantitative-ablation} and Figure \ref{fig:ablation_loss}: $ \mathcal{L}_{bg} $, $ \mathcal{L}_{id} $, and $ \mathcal{L}_{norm} $ all contribute to the maintenance of irrelevant attributes, and $ \mathcal{L}_{s\_mc} $ helps keep hair color unchanged when only editing the hairstyle.

\begin{table}[t]
	\centering
	\setlength{\tabcolsep}{0.8em}{
		\begin{tabular}{lcccc}
			\hline
			Methods & IDS & PSNR & SSIM & ACD\\
			\hline
			Ours  & \textbf{0.85} & \textbf{27.0} & \textbf{0.91} & \textbf{0.02} \\
			\hline
			w/o $ \mathcal{L}_{bg} $ & 0.82 & 19.9 & 0.82 & \textbf{0.02} \\
			w/o $ \mathcal{L}_{id} $ & 0.25 & 22.8 & 0.80 & 0.03 \\
			w/o $ \mathcal{L}_{s\_mc} $ & 0.82 & 26.6 & 0.90 & 0.09  \\
			w/o $ \mathcal{L}_{norm} $ & 0.75 & 24.9 & 0.87 & 0.03 \\
			\hline
		\end{tabular}
	}
	\caption{Quantitative ablation experiments on attribute preservation loss.}
	\label{tab:quantitative-ablation}
	\vspace{-1em}
\end{table}

\begin{figure}[tb]
	\begin{center}
		\setlength{\tabcolsep}{1pt}
		\begin{tabular}{ccccc}
			\small{Input Image} & \small{Ours} & \small{(a)}  & \small{(b)} & \small{(c)}
			\\
			\includegraphics[width=1.6cm]{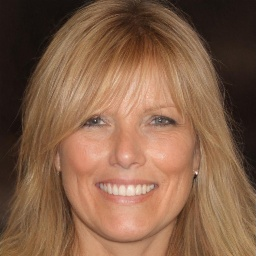}
			&\includegraphics[width=1.6cm]{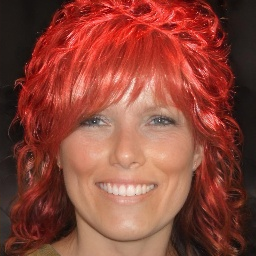}
			&\includegraphics[width=1.6cm]{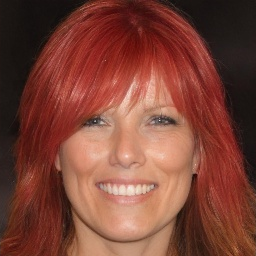}
			&\includegraphics[width=1.6cm]{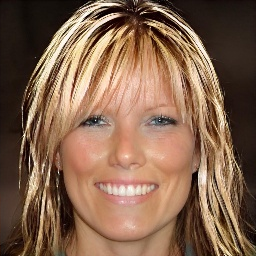}
			&\includegraphics[width=1.6cm]{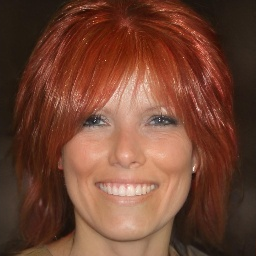}
			\\
			
			\includegraphics[width=1.6cm]{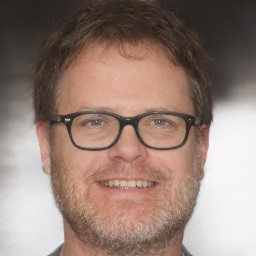}
			&\includegraphics[width=1.6cm]{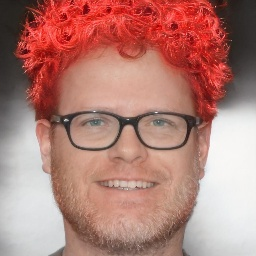}
			&\includegraphics[width=1.6cm]{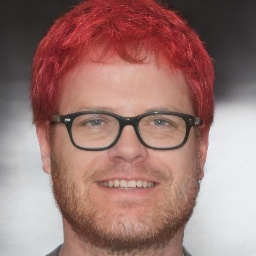}
			&\includegraphics[width=1.6cm]{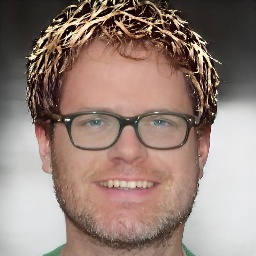}
			&\includegraphics[width=1.6cm]{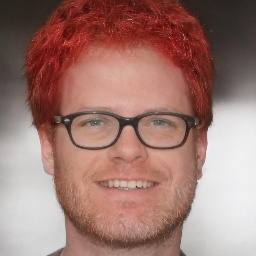}
			
			\\
		\end{tabular}
	\end{center}
	\vspace{-0.5em}
	\caption{Visual comparison of the results generated by our method and variants of our model. The text description is ``perm hairstyle and red hair". (a) concatenate conditional inputs with the latent code. (b) replace the conditional inputs of the coarse and medium sub hair mappers with hair color embedding, and the fine sub hair mapper with hairstyle embedding. (c) replace the conditional input of the medium sub hair mapper with the hair color embedding and leave the rest unchanged.} 
	\label{fig:ablation-network}
	\vspace{-1em}
\end{figure}

\noindent\textbf{Superiority of Network Structure Design.} We compare our model with three variants. (a) replace the modulation module with vanilla layernorm layer, and concatenate conditional inputs with the latent code and then feed them into the network. (b) replace the conditional inputs of the coarse and medium sub hair mappers with hair color embedding, and the fine sub hair mapper with hairstyle embedding. (c) replace the conditional input of the medium sub hair mapper with the hair color embedding and leave the rest unchanged. As shown in Figure \ref{fig:ablation-network}, only our model completes both hairstyle and hair color manipulation. The unsatisfactory result of (a) proves that our \textit{modulation module} can better fuse the condition information into latent space and improve manipulation capability. (b) and (c) confirm the correctness of our \textit{disentangled semantic-matching-based information injection}.

\begin{figure}[tb]
	\begin{center}
		\setlength{\tabcolsep}{0.5pt}
		\begin{tabular}{cccccc}
			\makecell[c]{ \\ Input} & \makecell[c]{ \\ Ours} & \makecell[c]{w/o \\ $ \mathcal{L}_{bg} $} & \makecell[c]{w/o \\ $ \mathcal{L}_{id} $} & \makecell[c]{w/o \\ $ \mathcal{L}_{s\_mc} $} & \makecell[c]{w/o \\ $ \mathcal{L}_{norm} $}
			\\
			\includegraphics[width=1.34cm]{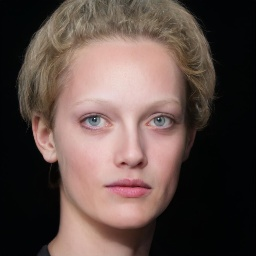}
			&\includegraphics[width=1.34cm]{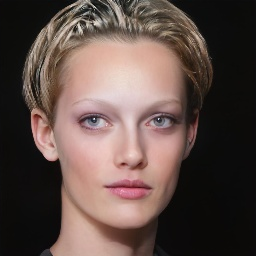}
			&\includegraphics[width=1.34cm]{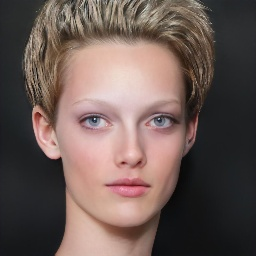}
			&\includegraphics[width=1.34cm]{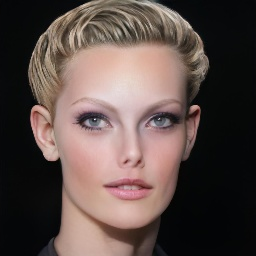}
			&\includegraphics[width=1.34cm]{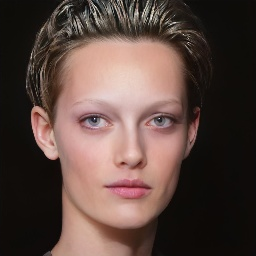}
			&\includegraphics[width=1.34cm]{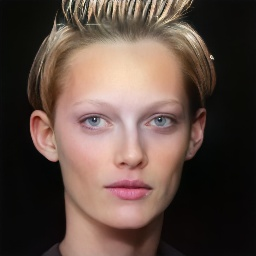}
			\\
			
			\includegraphics[width=1.34cm]{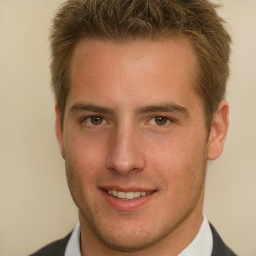}
			&\includegraphics[width=1.34cm]{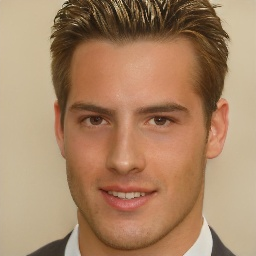}
			&\includegraphics[width=1.34cm]{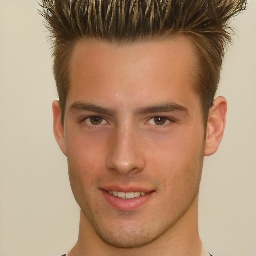}
			&\includegraphics[width=1.34cm]{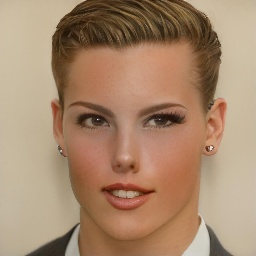}
			&\includegraphics[width=1.34cm]{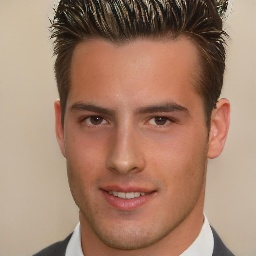}
			&\includegraphics[width=1.34cm]{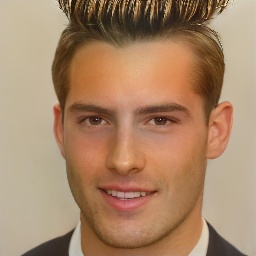}
			\\
			
		\end{tabular}
	\end{center}
	\vspace{-1em}
	\caption{The effect of attribute preservation loss. The text description is ``slicked back hairstyle".} 
	\vspace{-1em}
	\label{fig:ablation_loss}
\end{figure}

\begin{figure}[t]
	\centering
	\includegraphics[width=\columnwidth]{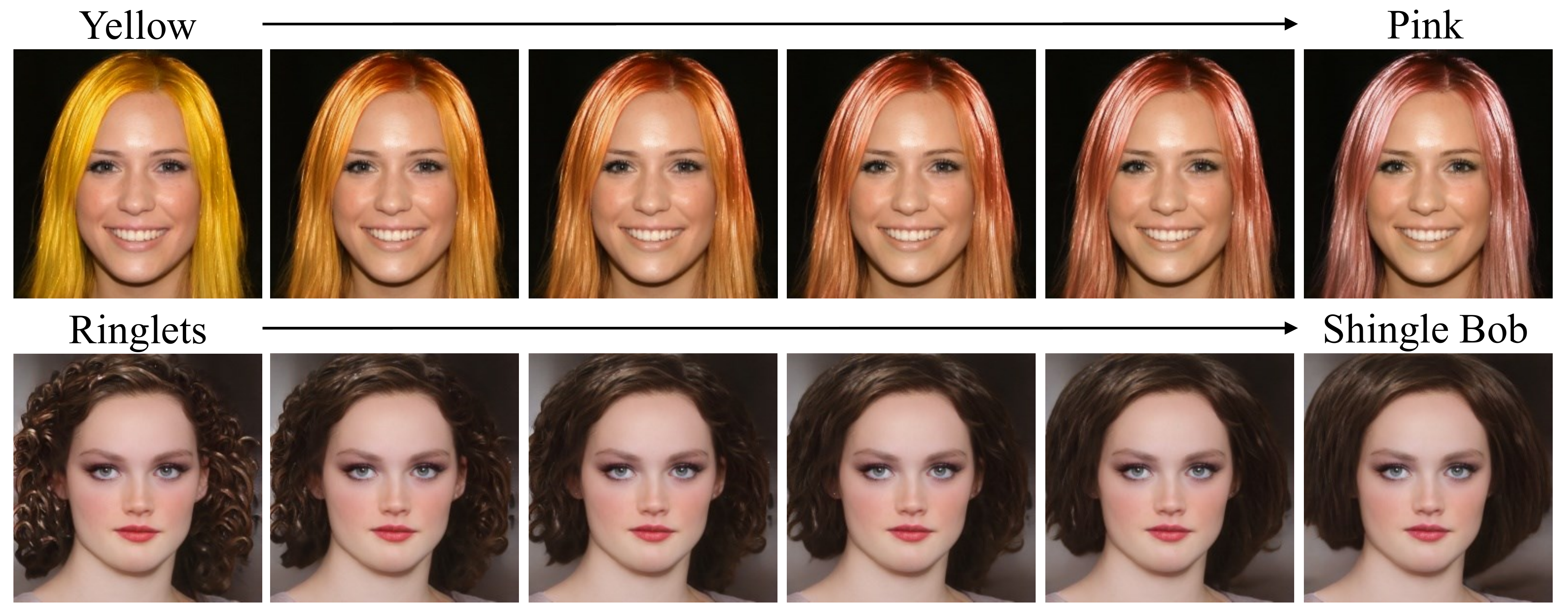} 
	\caption{Hair interpolation results. By gradually increasing the blending parameter $ \lambda $ from $ 0 $ to $ 1 $, we can manage hair editing at a fine-grained level, such as changing from yellow hair to pink hair, from ringlets hairstyle to shingle bob hairstyle.} 
	\label{fig:interpolation}
	\vspace{-1em}
\end{figure}

\noindent\textbf{Hair Interpolation.} Given two edited latent codes $ W_{A},W_{B} \in \mathcal{W+} $, we can achieve fine-grained hair editing by interpolation. In detail, we combine the two latent codes by linear weighting to generate the intermediate latent code $ W_{I}=\lambda W_{B}+(1-\lambda)W_{A}$. Finally, the image corresponding to the intermediate latent code is generated. By gradually increasing the blending parameter $ \lambda $ from $ 0 $ to $ 1 $, we can manage hair editing at a fine-grained level, as shown in Figure \ref{fig:interpolation}.

\begin{figure}[tb]
	\begin{center}
		\setlength{\tabcolsep}{1pt}
		\begin{tabular}{ccccc}
			\small{Input Image} & \small{Curly Short} & \small{Mushroom}  & \small{Violet} & \small{Silver}
			\\
			\includegraphics[width=1.6cm]{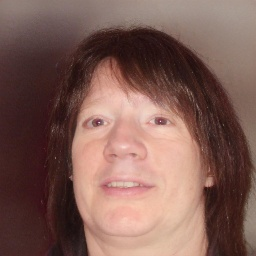}
			&\includegraphics[width=1.6cm]{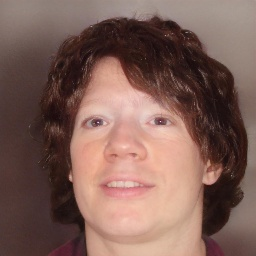}
			&\includegraphics[width=1.6cm]{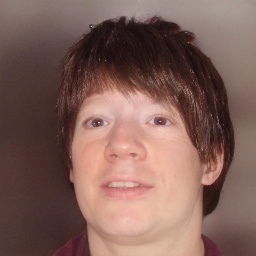}
			&\includegraphics[width=1.6cm]{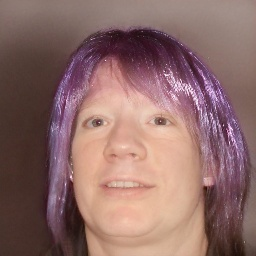}
			&\includegraphics[width=1.6cm]{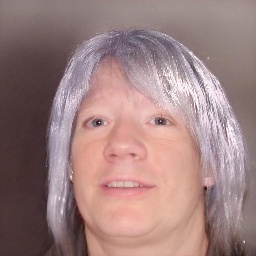}
			\\
			
			\includegraphics[width=1.6cm]{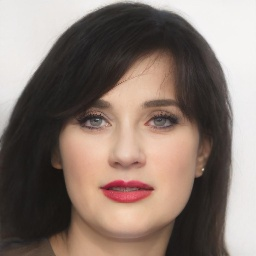}
			&\includegraphics[width=1.6cm]{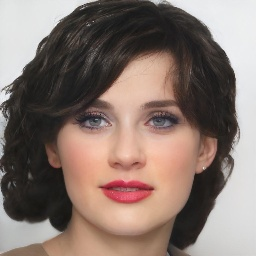}
			&\includegraphics[width=1.6cm]{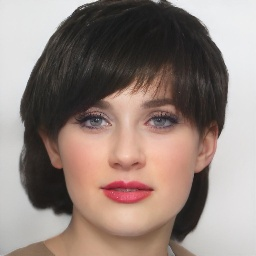}
			&\includegraphics[width=1.6cm]{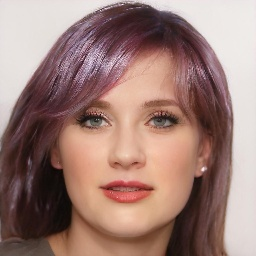}
			&\includegraphics[width=1.6cm]{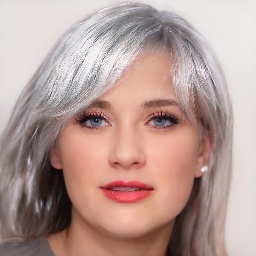}
			\\
		\end{tabular}
	\end{center}
	\vspace{-0.5em}
	\caption{Generalization ability to unseen descriptions. Despite never being trained on these descriptions of ``curly short hairstyle", ``mushroom hairstyle", ``violet hair", and ``silver hair", our method can still yield plausible manipulation results.} 
	\label{fig:extension}
\end{figure}

\noindent\textbf{Generalization Ability.} In Figure \ref{fig:extension}, we demonstrate the generalization ability of our method to unseen text descriptions. Thanks to our strategy of \textit{shared condition embedding}, our method possesses some extrapolation ability after training with only a limited number of hair editing descriptions, which yields reasonable editing results for texts that never appear in the training descriptions.

\noindent\textbf{Cross-Modal Conditional Inputs.} Our method supports conditional inputs from the image and text domains individually or jointly, which is not feasible with current existing hair editing methods, and the results are shown in Figure \ref{fig:teaser}. More results will be given in the supplementary materials.

\section{Limitations and Negative Impact}

Since our editing is done in the latent space of pretrained StyleGAN, we can not complete the editing for some rare hairstyle descriptions or reference images that are not within the domain of StyleGAN. But this limitation can be well solved by adding corresponding images to the pre-training process of StyleGAN. In the hairstyle transfer task, we use the embedding of the hairstyle reference image in the latent space of CLIP as the conditional input for our hair mapper, which sometimes will lose the fine-grained structural information and thus cannot achieve a perfect transfer for the hairstyle's structural details. In addition, the hair-edited images by our method may be used to spread malicious information, which can be evaded using the existing state-of-the-art GAN-generated image detectors~\cite{wang2020cnn}.

\section{Conclusions}
In this paper, we propose a new hair editing interaction mode that unifies conditional inputs from text and image domains in a unified framework. In our framework, users can individually or jointly provide textual descriptions and reference images to complete the hair editing. This multi-modal interaction greatly increases the flexibility of hair editing and reduces the interaction cost for users. By maximizing the great potential of CLIP, tailored network structure designs and loss functions, our framework supports high-quality hair editing in a decoupled manner. Extensive qualitative and quantitative comparisons and user study demonstrate the superiority of our method compared to competing methods in terms of manipulation capability, irrelevant attributes preservation, and image realism.

%%%%%%%%% REFERENCES
{\small
\bibliographystyle{ieee_fullname}
\bibliography{egbib}
}

\clearpage
\appendix
%\appendixpage

\twocolumn[{
	\renewcommand\twocolumn[1][]{#1}
%	\maketitle
	
	%\begin{table*}[t]
	\centering
	\small
	\setlength{\tabcolsep}{0.21em}{
		\begin{tabular}{l | cccc | cccc | cccc | cccc | cccc}
			\hline
			\multicolumn{1}{ c }{{}}	& \multicolumn{4}{ c }{{Afro Hairstyle}} & \multicolumn{4}{ c }{{Bobcut Hairstyle}} & \multicolumn{4}{ c }{{Bowlcut Hairstyle}} & \multicolumn{4}{ c }{{Mohawk Hairstyle}} & \multicolumn{4}{ c }{{Purple Hair}} \\
			\hline
			Methods & FID & IDS & PSNR & SSIM & FID & IDS & PSNR & SSIM & FID & IDS & PSNR & SSIM & FID & IDS & PSNR & SSIM & FID & IDS & PSNR & SSIM \\
			\hline
			Ours  & 40.9 & 0.84 & 27.8 & 0.92  & 15.2 & 0.85 & 28.1 & 0.92  & 44.4 & 0.78 & 26.8 & 0.90 & 39.7 & 0.82 & 26.4 & 0.89  & 62.2 & 0.90 & 30.8 & 0.95 \\
			StyleCLIP  & 73.1 & 0.79 & 24.5 & 0.87  & 29.2 & 0.77 & 24.7 & 0.88  & 48.2 & 0.74 & 24.2 & 0.88  & 38.5 & 0.70 & 20.6 & 0.81  & 83.4 & 0.87 & 24.3 & 0.90 \\
			TediGAN & 29.0 & 0.17 & 24.2 & 0.79  & 26.4 & 0.15 & 23.9 & 0.79  & 29.4 & 0.14 & 23.8 & 0.79  & 27.4 & 0.17 & 24.2 & 0.79  & 26.7 & 0.18 & 24.5 & 0.80 \\
			\hline
			
			\multicolumn{1}{ c }{{}}	& \multicolumn{4}{ c }{{Green Hair}} & \multicolumn{4}{ c }{{Blond Hair}} & \multicolumn{4}{ c }{{Braid Brown Hair}} & \multicolumn{4}{ c }{{Crewcut Yellow Hair}} & \multicolumn{4}{ c }{{Perm Gray Hair}} \\
			\hline
			Methods & FID & IDS & PSNR & SSIM & FID & IDS & PSNR & SSIM & FID & IDS & PSNR & SSIM & FID & IDS & PSNR & SSIM & FID & IDS & PSNR & SSIM\\
			\hline
			Ours  & 62.1 & 0.90 & 30.2 & 0.95  & 23.1 & 0.87 & 30.0 & 0.95  & 34.2 & 0.76 & 26.2 & 0.89  & 69.5 & 0.78 & 25.8 & 0.89  & 93.4 & 0.82 & 26.3 & 0.91 \\
			StyleCLIP  & 81.2 & 0.85 & 22.8 & 0.87  & 53.5 & 0.81 & 26.3 & 0.91  & 22.3 & 0.81 & 23.3 & 0.89  & 84.0 & 0.78 & 20.1 & 0.83  & 76.6 & 0.80 & 20.8 & 0.85 \\
			TediGAN & 34.7 & 0.17 & 24.3 & 0.79  & 30.3 & 0.18 & 24.4 & 0.80  & 29.9 & 0.16 & 23.9 & 0.79  & 32.0 & 0.18 & 24.3 & 0.80  & 33.2 & 0.15 & 23.9 & 0.79 \\
			\hline			
			
		\end{tabular}
	}
	\captionof{table}{Quantitative comparisons regarding FID \cite{Heusel2017GANsTB} and the preservation of irrelevant attributes. FID measures the distributions of images before and after editing. IDS denotes identity similarity before and after editing. PSNR and SSIM are calculated in the region of intersection of non-hair regions before and after editing. Our approach exhibits the best irrelevant attributes preservation ability.}
	\vspace{1.5em}
	\label{tab:textcomparetable-supp}
}]	
%\end{table*}

\section{Quantitative Results}

We give the detailed quantitative results about FID \cite{Heusel2017GANsTB} in Table \ref{tab:textcomparetable-supp} for each hair description. Although TediGAN \cite{xia2021tedigan} performs the best in terms of FID, it hardly performs any hair editing task very well (as demonstrated by the qualitative results). We argue that FID may not be a suitable metric for evaluating the manipulation ability. Similar conclusion was also drawn in e4e~\cite{Tov2021DesigningAE}.

\section{More Qualitative Results}
\begin{figure*}[tb]
	\begin{center}
		\setlength{\tabcolsep}{0.5pt}
		
		\begin{tabular}{m{0.3cm}<{\centering}m{2.05cm}<{\centering}m{2.05cm}<{\centering}m{2.05cm}<{\centering}m{2.05cm}<{\centering}m{2.05cm}<{\centering}m{2.05cm}<{\centering}m{2.05cm}<{\centering}m{2.05cm}<{\centering}}
			& \small{Input Image} & \small{Ours} & \small{StyleCLIP~\cite{patashnik2021styleclip}} & \small{TediGAN~\cite{xia2021tedigan}} & \small{Input Image} & \small{Ours} & \small{StyleCLIP~\cite{patashnik2021styleclip}} & \small{TediGAN~\cite{xia2021tedigan}}
			\\
			
			\raisebox{0.7cm}{\rotatebox[origin=c]{90}{\footnotesize{{afro hairstyle}}}}
			&\includegraphics[width=2cm]{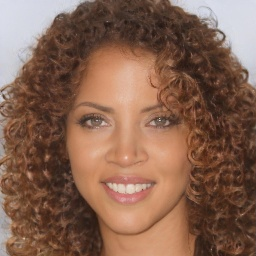}
			&\includegraphics[width=2cm]{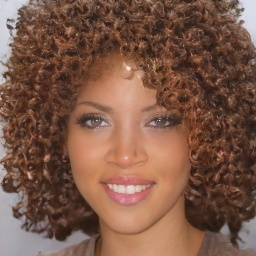}
			&\includegraphics[width=2cm]{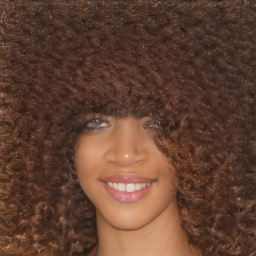}
			&\includegraphics[width=2cm]{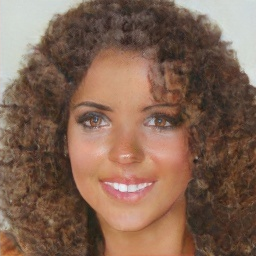}
			&\includegraphics[width=2cm]{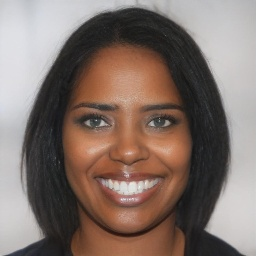}
			&\includegraphics[width=2cm]{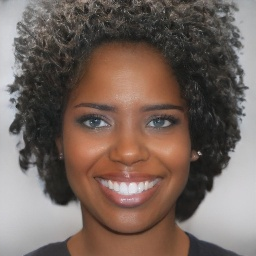}
			&\includegraphics[width=2cm]{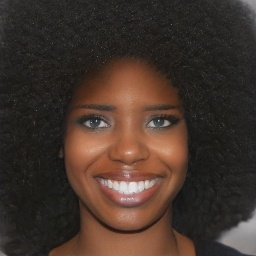}
			&\includegraphics[width=2cm]{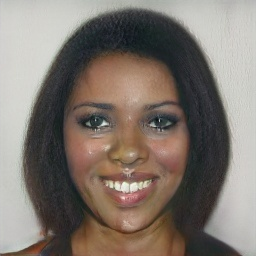}
			\\
			
			\raisebox{0.83cm}{\rotatebox[origin=c]{90}{\footnotesize{{bobcut hairstyle}}}}
			&\includegraphics[width=2cm]{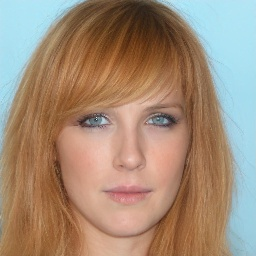}
			&\includegraphics[width=2cm]{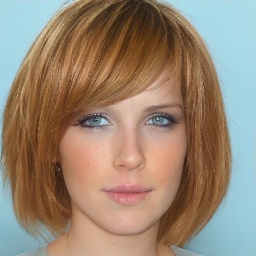}
			&\includegraphics[width=2cm]{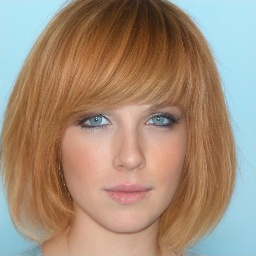}
			&\includegraphics[width=2cm]{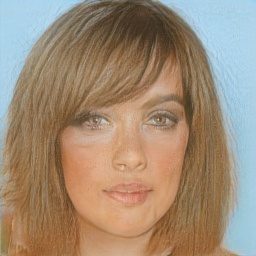}
			&\includegraphics[width=2cm]{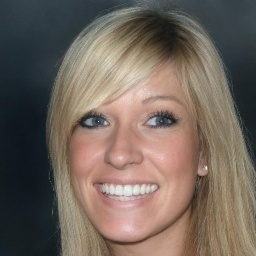}
			&\includegraphics[width=2cm]{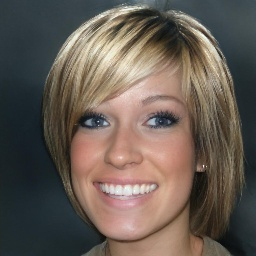}
			&\includegraphics[width=2cm]{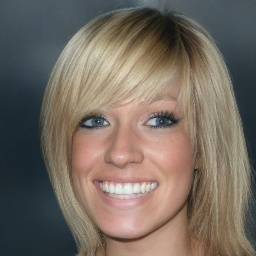}
			&\includegraphics[width=2cm]{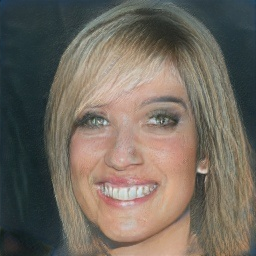}
			\\
			
			\raisebox{0.87cm}{\rotatebox[origin=c]{90}{\footnotesize{{bowlcut hairstyle}}}}
			&\includegraphics[width=2cm]{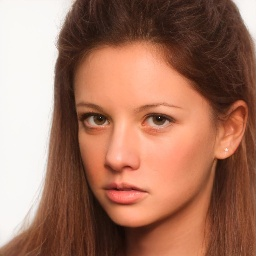}
			&\includegraphics[width=2cm]{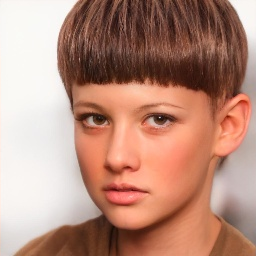}
			&\includegraphics[width=2cm]{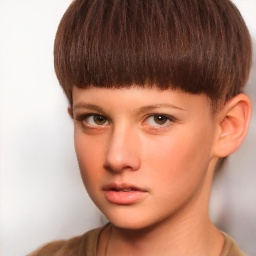}
			&\includegraphics[width=2cm]{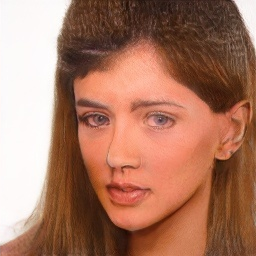}
			&\includegraphics[width=2cm]{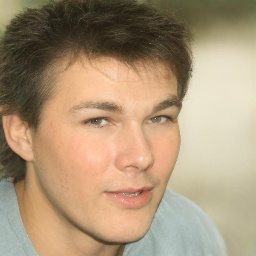}
			&\includegraphics[width=2cm]{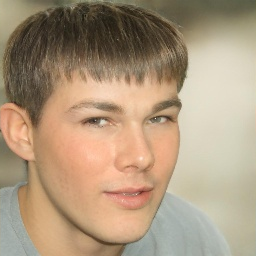}
			&\includegraphics[width=2cm]{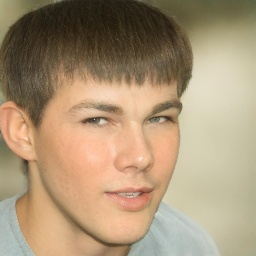}
			&\includegraphics[width=2cm]{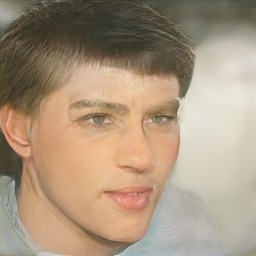}
			\\
			
			\raisebox{0.93cm}{\rotatebox[origin=c]{90}{\footnotesize{{mohawk hairstyle}}}}
			&\includegraphics[width=2cm]{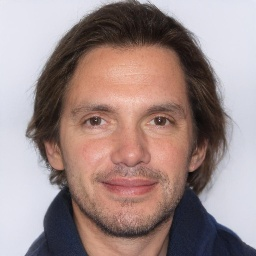}
			&\includegraphics[width=2cm]{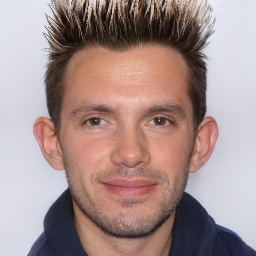}
			&\includegraphics[width=2cm]{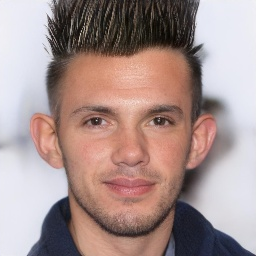}
			&\includegraphics[width=2cm]{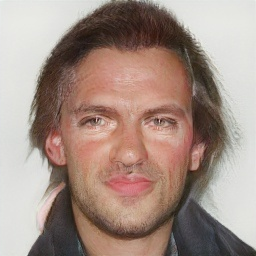}
			&\includegraphics[width=2cm]{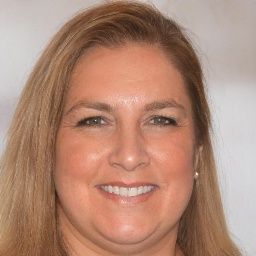}
			&\includegraphics[width=2cm]{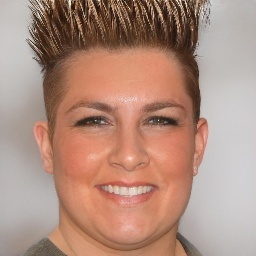}
			&\includegraphics[width=2cm]{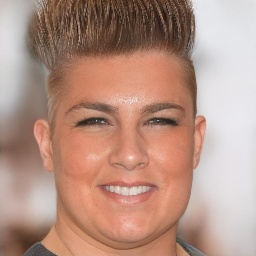}
			&\includegraphics[width=2cm]{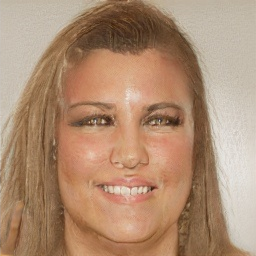}
			\\
			
			\raisebox{0.51cm}{\rotatebox[origin=c]{90}{\footnotesize{{purple hair}}}}
			&\includegraphics[width=2cm]{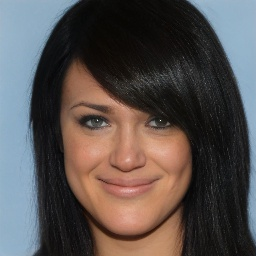}
			&\includegraphics[width=2cm]{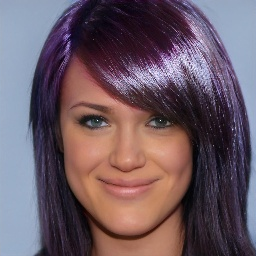}
			&\includegraphics[width=2cm]{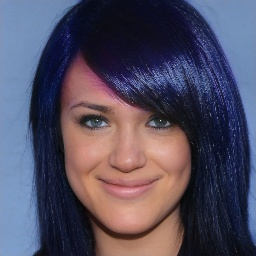}
			&\includegraphics[width=2cm]{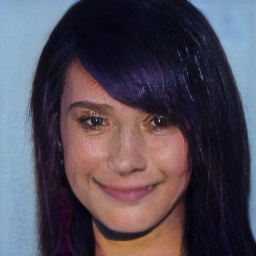}
			&\includegraphics[width=2cm]{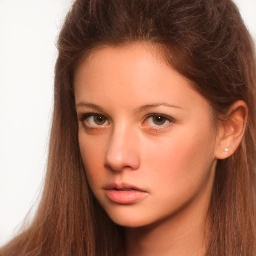}
			&\includegraphics[width=2cm]{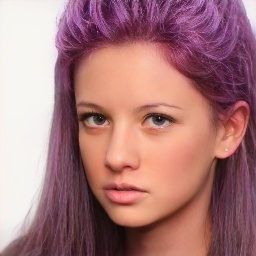}
			&\includegraphics[width=2cm]{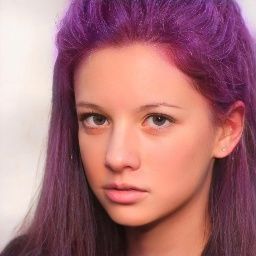}
			&\includegraphics[width=2cm]{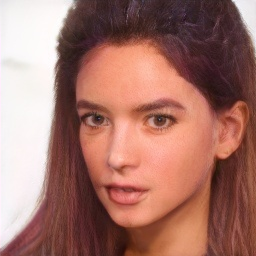}
			\\

			\raisebox{0.51cm}{\rotatebox[origin=c]{90}{\footnotesize{{green hair}}}}
			&\includegraphics[width=2cm]{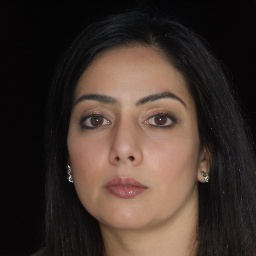}
			&\includegraphics[width=2cm]{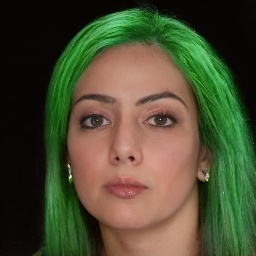}
			&\includegraphics[width=2cm]{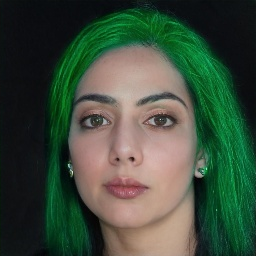}
			&\includegraphics[width=2cm]{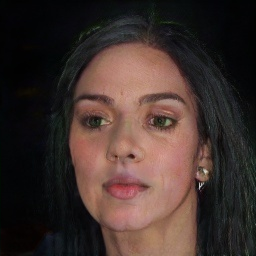}
			&\includegraphics[width=2cm]{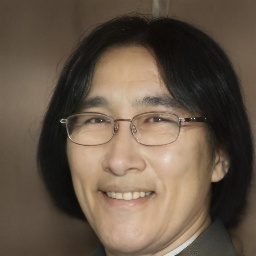}
			&\includegraphics[width=2cm]{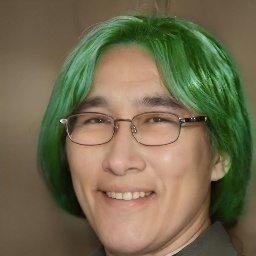}
			&\includegraphics[width=2cm]{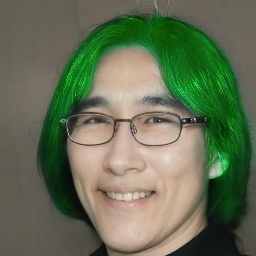}
			&\includegraphics[width=2cm]{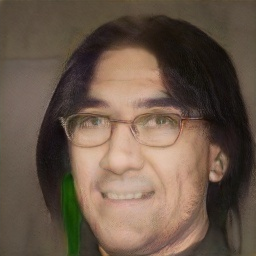}	
			\\
			
			\raisebox{0.45cm}{\rotatebox[origin=c]{90}{\footnotesize{{blond hair}}}}
			&\includegraphics[width=2cm]{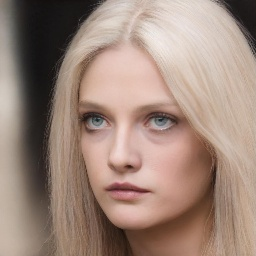}
			&\includegraphics[width=2cm]{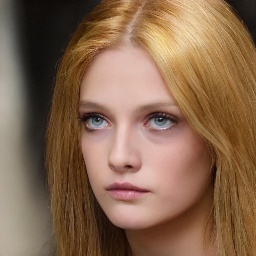}
			&\includegraphics[width=2cm]{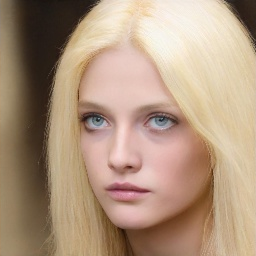}
			&\includegraphics[width=2cm]{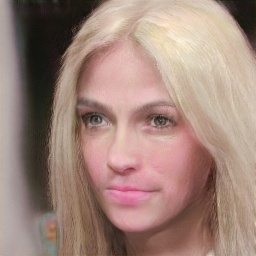}	
			&\includegraphics[width=2cm]{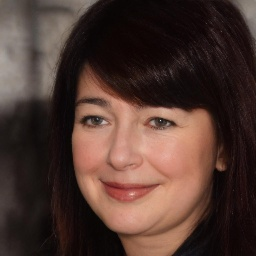}
			&\includegraphics[width=2cm]{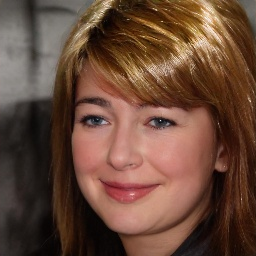}
			&\includegraphics[width=2cm]{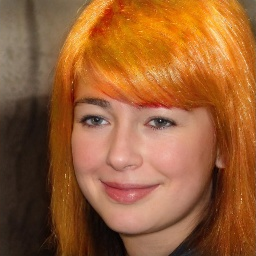}
			&\includegraphics[width=2cm]{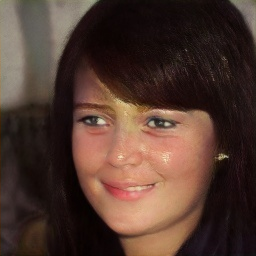}	
			\\
			
			\raisebox{0.51cm}{\rotatebox[origin=c]{90}{\footnotesize{{braid brown}}}}
			&\includegraphics[width=2cm]{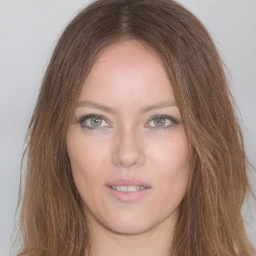}
			&\includegraphics[width=2cm]{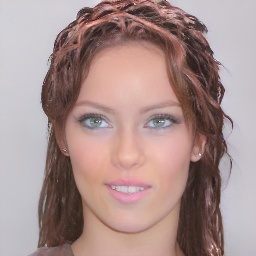}
			&\includegraphics[width=2cm]{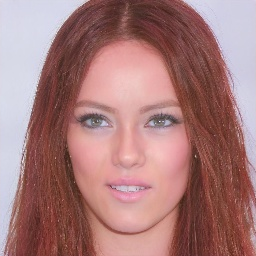}
			&\includegraphics[width=2cm]{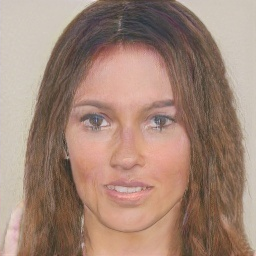}
			&\includegraphics[width=2cm]{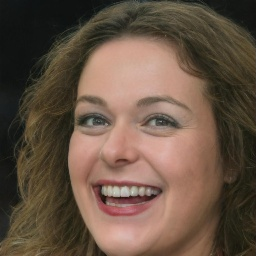}
			&\includegraphics[width=2cm]{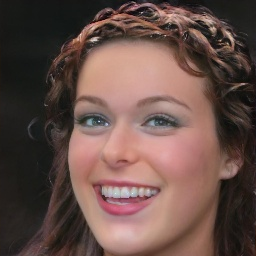}
			&\includegraphics[width=2cm]{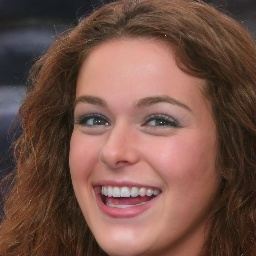}
			&\includegraphics[width=2cm]{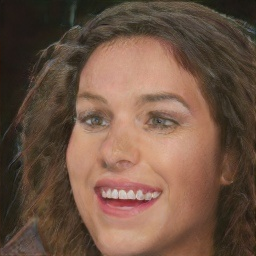}
			\\
			
			\raisebox{0.75cm}{\rotatebox[origin=c]{90}{\footnotesize{{crewcut yellow}}}}
			&\includegraphics[width=2cm]{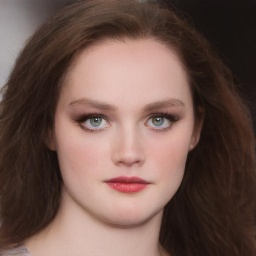}
			&\includegraphics[width=2cm]{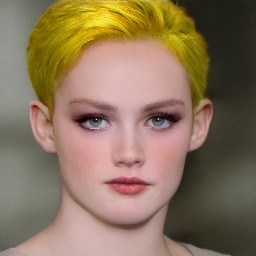}
			&\includegraphics[width=2cm]{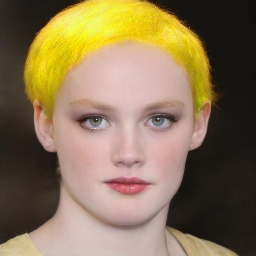}
			&\includegraphics[width=2cm]{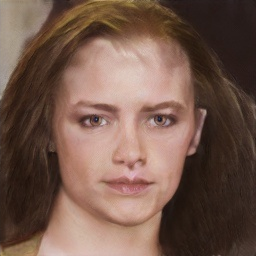}
			&\includegraphics[width=2cm]{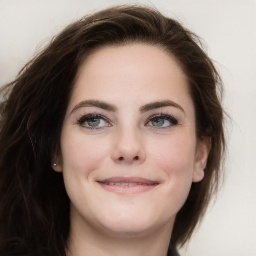}
			&\includegraphics[width=2cm]{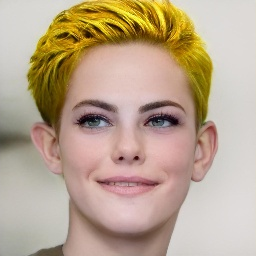}
			&\includegraphics[width=2cm]{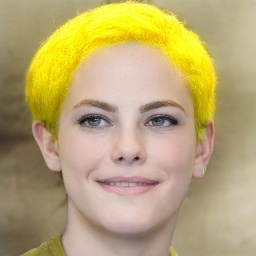}
			&\includegraphics[width=2cm]{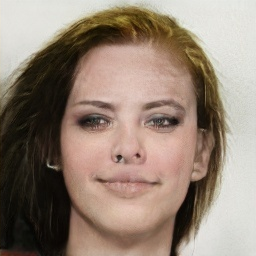}
			\\
			
			\raisebox{0.45cm}{\rotatebox[origin=c]{90}{\footnotesize{{perm gray}}}}
			&\includegraphics[width=2cm]{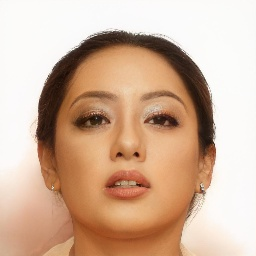}
			&\includegraphics[width=2cm]{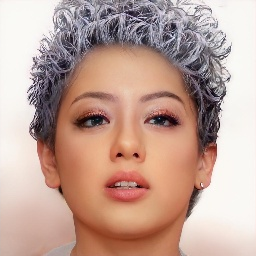}
			&\includegraphics[width=2cm]{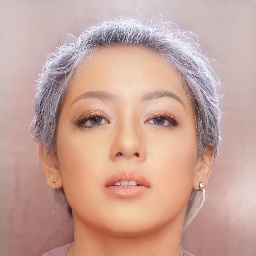}
			&\includegraphics[width=2cm]{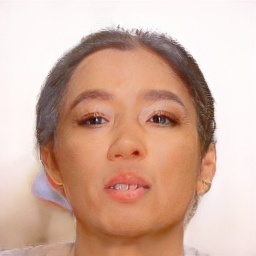}
			&\includegraphics[width=2cm]{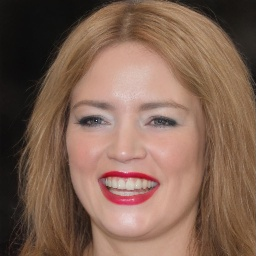}
			&\includegraphics[width=2cm]{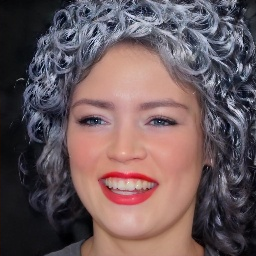}
			&\includegraphics[width=2cm]{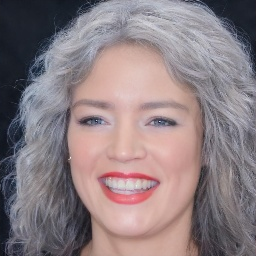}
			&\includegraphics[width=2cm]{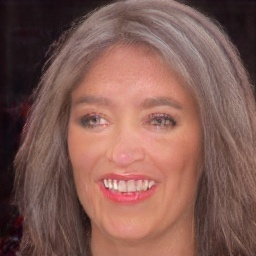}
			\\			
			
		\end{tabular}
	\end{center}
	\vspace{-1em}
	\caption{Visual comparison with StyleCLIP~\cite{patashnik2021styleclip} and TediGAN~\cite{xia2021tedigan}. The corresponding simplified text descriptions (editing hairstyle, hair color, or both of them) are listed on the leftmost side of each row, and all input images are the inversions of the real images. Our approach demonstrates better visual photorealism and irrelevant attributes preservation ability while completing the specified hair editing.} 
	\label{fig:textcomparefig-supp}
\end{figure*}

\begin{figure*}[t]
	\begin{center}
		\setlength{\tabcolsep}{1pt}
		\begin{tabular}{cccccc}
			Input & Hairstyle Ref & Color Ref & Ours & LOHO~\cite{Saha2021LOHOLO} & MichiGAN~\cite{tan2020michigan}
			\\
			\includegraphics[width=2.8cm]{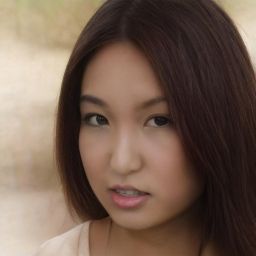}
			&\includegraphics[width=2.8cm]{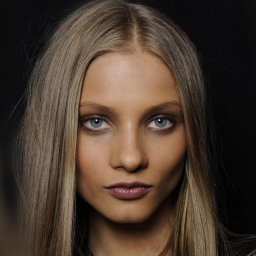}
			&\includegraphics[width=2.8cm]{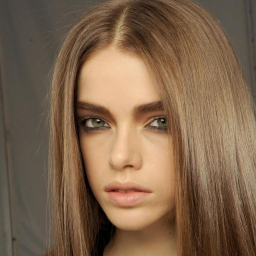}
			&\includegraphics[width=2.8cm]{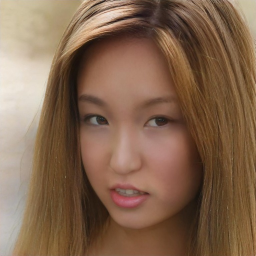}
			&\includegraphics[width=2.8cm]{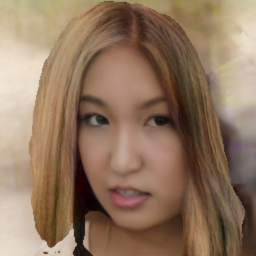}
			&\includegraphics[width=2.8cm]{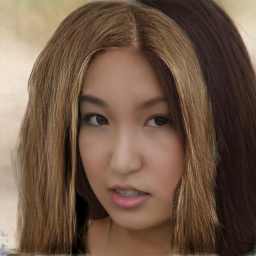}
			\\
			
			\includegraphics[width=2.8cm]{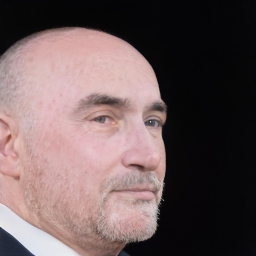}
			&\includegraphics[width=2.8cm]{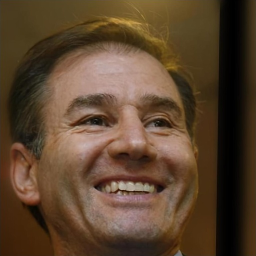}
			&\includegraphics[width=2.8cm]{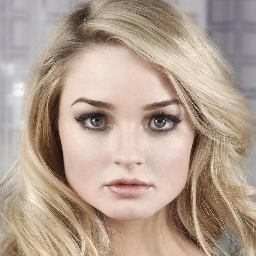}
			&\includegraphics[width=2.8cm]{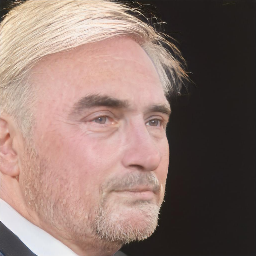}
			&\includegraphics[width=2.8cm]{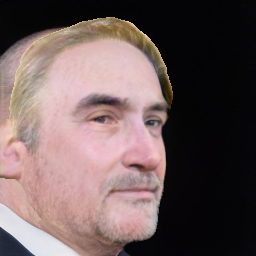}
			&\includegraphics[width=2.8cm]{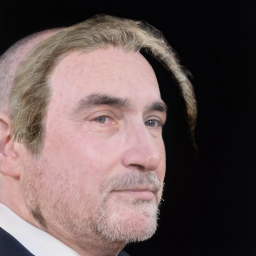}
			\\
			
			\includegraphics[width=2.8cm]{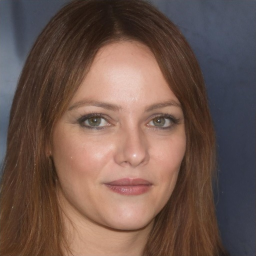}
			&\includegraphics[width=2.8cm]{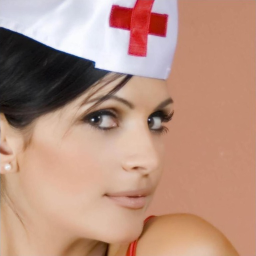}
			&\includegraphics[width=2.8cm]{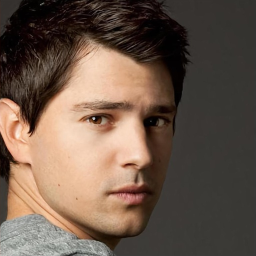}
			&\includegraphics[width=2.8cm]{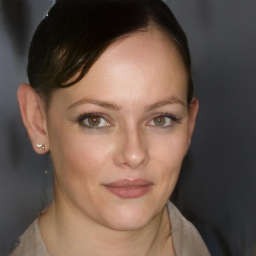}
			&\includegraphics[width=2.8cm]{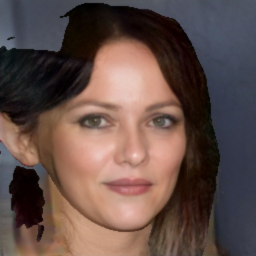}
			&\includegraphics[width=2.8cm]{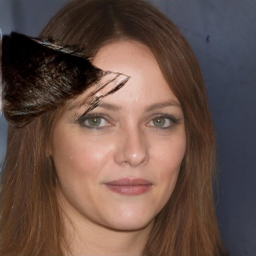}
			\\
			
			\includegraphics[width=2.8cm]{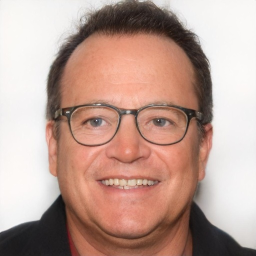}
			&\includegraphics[width=2.8cm]{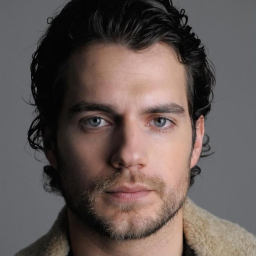}
			&\includegraphics[width=2.8cm]{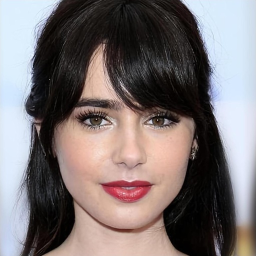}
			&\includegraphics[width=2.8cm]{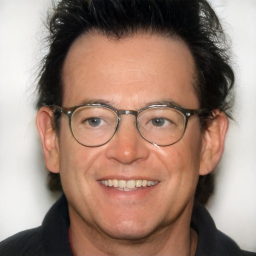}
			&\includegraphics[width=2.8cm]{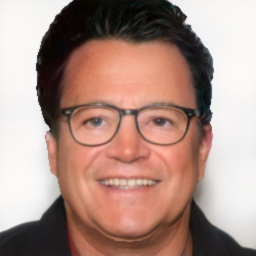}
			&\includegraphics[width=2.8cm]{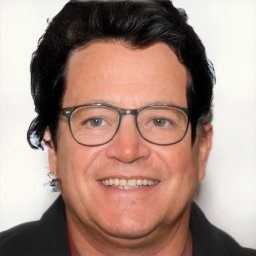}
			\\			
			
			\includegraphics[width=2.8cm]{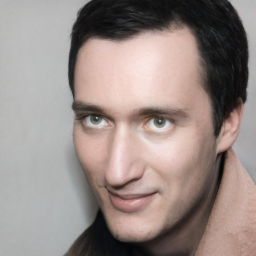}
			&\includegraphics[width=2.8cm]{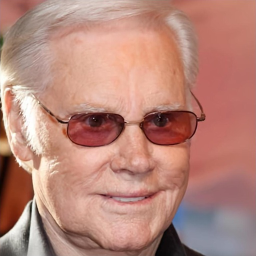}
			&\includegraphics[width=2.8cm]{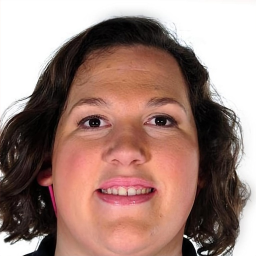}
			&\includegraphics[width=2.8cm]{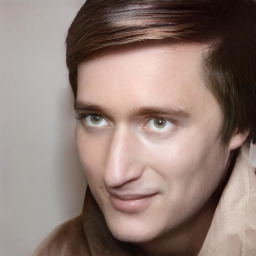}
			&\includegraphics[width=2.8cm]{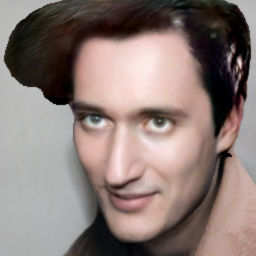}
			&\includegraphics[width=2.8cm]{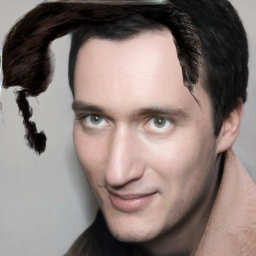}
			\\	
			
			\includegraphics[width=2.8cm]{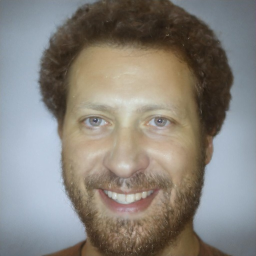}
			&\includegraphics[width=2.8cm]{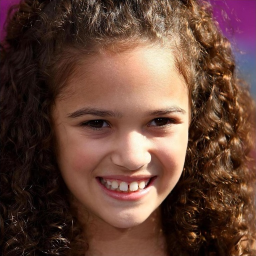}
			&\includegraphics[width=2.8cm]{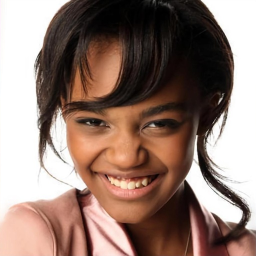}
			&\includegraphics[width=2.8cm]{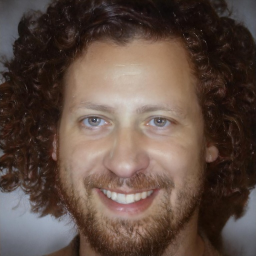}
			&\includegraphics[width=2.8cm]{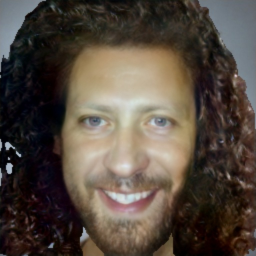}
			&\includegraphics[width=2.8cm]{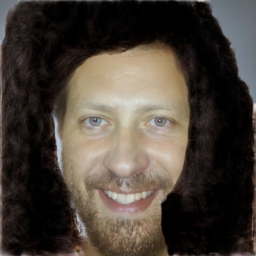}
			\\
			
			\includegraphics[width=2.8cm]{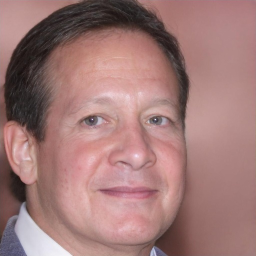}
			&\includegraphics[width=2.8cm]{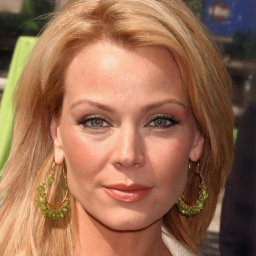}
			&\includegraphics[width=2.8cm]{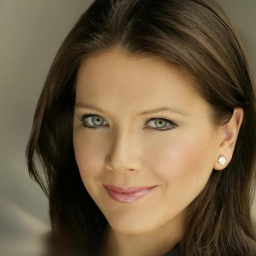}
			&\includegraphics[width=2.8cm]{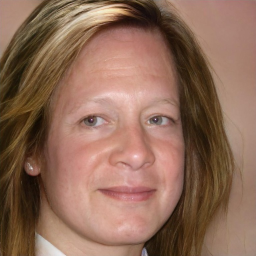}
			&\includegraphics[width=2.8cm]{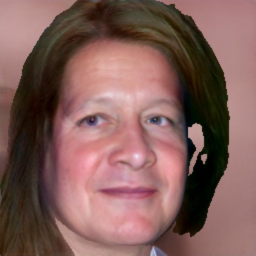}
			&\includegraphics[width=2.8cm]{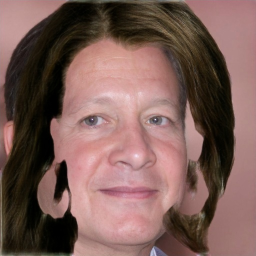}
			\\			
		\end{tabular}
	\end{center}
	\caption{Comparison of our approach with LOHO~\cite{Saha2021LOHOLO} and MichiGAN~\cite{tan2020michigan} on hair transfer. Even for extreme examples like the third row in this figure, our method can yield plausible hairstyle transfer results.} 
	\label{fig:transfer-compare-supp}
	
\end{figure*}

\begin{figure*}[t]
	\centering
	\includegraphics[width=\textwidth]{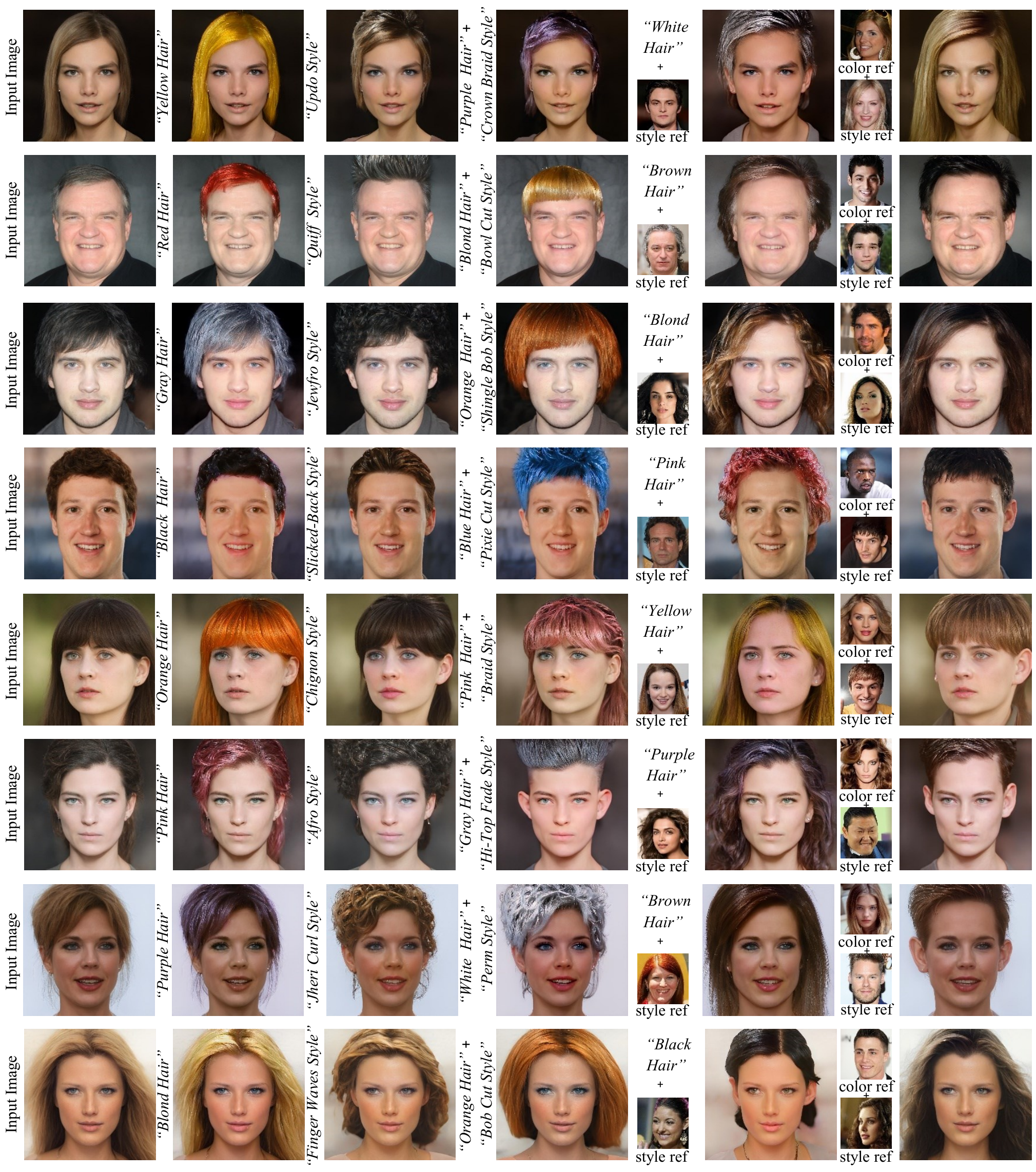} 
	\caption{Our single framework supports hairstyle and hair color editing individually or jointly, and conditional inputs can come from either image or text domain. ``Style'' in the text description is the abbreviation for hairstyle.} 
	\label{fig:crossmodal}
\end{figure*}

In Figures \ref{fig:textcomparefig-supp}, \ref{fig:transfer-compare-supp}, and \ref{fig:crossmodal} we give more visual comparison results with other state-of-the-art methods and results for the cross-modal conditional inputs.

\end{document}